\documentclass[letterpaper]{article} 
\usepackage[submission]{aaai23}  
\usepackage{times}  
\usepackage{helvet}  
\usepackage{courier}  
\usepackage[hyphens]{url}  
\usepackage{graphicx} 
\usepackage{natbib}  
\usepackage{caption} 
\usepackage{algorithm}
\usepackage{algorithmic}
\newcommand{\bb}[1]{\textbf{#1}}
\usepackage{amsmath,amssymb}
\usepackage{newfloat}
\usepackage{listings}
\DeclareCaptionStyle{ruled}{labelfont=normalfont,labelsep=colon,strut=off} 
\floatstyle{ruled}
\newfloat{listing}{tb}{lst}{}
\floatname{listing}{Listing}
\title{RGBD1K: A Large-scale Dataset and Benchmark for RGB-D Object Tracking}
\author{
    Xue-Feng Zhu\textsuperscript{\rm 1},
    Tianyang Xu\textsuperscript{\rm 1},
    Zhangyong Tang\textsuperscript{\rm 1},
    Zucheng Wu\textsuperscript{\rm 1},\\
    Haodong Liu\textsuperscript{\rm 1},
    Xiao Yang\textsuperscript{\rm 1},
    Xiao-Jun Wu\textsuperscript{\rm 1}\thanks{This author is the corresponding author.},
    Josef Kittler\textsuperscript{\rm 2}
}
\affiliations{
    \textsuperscript{\rm 1}School of Artificial Intelligence and Computer Science, Jiangnan University, Wuxi, Jiangsu, P.R. China.\\
    \textsuperscript{\rm 2}Centre for Vision, Speech and Signal Processing, University of Surrey, Guildford, GU2 7XH, UK\\
    
    \textit{Email: xuefeng\_zhu95@163.com; \{tianyang.xu; wu\_xiaojun\}@jiangnan.edu.cn; j.kittler@surrey.ac.uk}
    
}

\usepackage{bibentry}

\begin{document}

\maketitle

\begin{abstract}
RGB-D object tracking has attracted considerable attention recently, achieving promising performance thanks to the symbiosis between visual and depth channels.
However, given a limited amount of annotated RGB-D tracking data, most state-of-the-art RGB-D trackers are simple extensions of high-performance RGB-only trackers, without fully exploiting the underlying potential of the depth channel in the offline training stage.
To address the dataset deficiency issue, a new RGB-D dataset named RGBD1K is released in this paper.
The RGBD1K contains 1,050 sequences with about 2.5M frames in total. To demonstrate the benefits of training on a larger RGB-D data set in general, and RGBD1K in particular, we develop a transformer-based RGB-D tracker, named SPT, as a baseline for future visual object tracking studies using the new dataset. The results, of extensive experiments using the SPT tracker emonstrate the potential of the RGBD1K dataset to improve the performance of RGB-D tracking, inspiring future developments of effective tracker designs.
The dataset and codes will be available on the project homepage:~\url{https://github.com/xuefeng-zhu5/RGBD1K}.
\end{abstract}

\section{Introduction}
Visual Object Tracking (VOT) aims at detecting the position and scale of an object of interest in every frame of a video. The tracking capability plays a significant role in the gamut of perceptual functionalities in computer vision and pattern recognition~\cite{xue2020semantic, griffiths2017empirical, smeulders2013visual}.
The development of visual object tracking techniques has been ongoing for decades.
In recent years in particular, with the access to large-scale annotated datasets, such as GOT10K~\cite{huang2019got}, TrackingNet~\cite{muller2018trackingnet}, LaSOT~\cite{fan2019lasot}, etc., the development of advanced visual object trackers has been accelerated by deep learning. 
Trained offline, using millions of labelled video frames, tracking networks are capable to learn robust feature representations, resulting in remarkable performance improvements, compared with conventional online learning methods~\cite{kristan2019seventh, kristan2020eighth, kristan2021ninth}. 

Recently, with the widespread availability of low-cost RGB-D sensors, the task of visual object tracking has broadened to include RGB-D videos. 
An RGB-D data is comprised of a three-channel RGB image and a single-channel depth map.
Compared to conventional RGB-only tracking, the additional depth maps of RGB-D videos provide supplementary spatial information that facilitates object tracking in complicated scenarios~\cite{bagautdinov2015probability, meshgi2016occlusion}.
Nevertheless, the existing RGB-D tracking methods generally build upon high-performance RGB-only trackers, adopting the depth information in the online tracking stage to support reasoning about partially occluded targets, and re-detection of disappearing targets~\cite{camplani2015real, kart2018make, hannuna2019ds}.

However, RGB-D trackers are not evolving as swiftly as RGB-only trackers~\cite{kristan2019seventh, kristan2020eighth, kristan2021ninth}.
The main reason is the lack of training data for RGB-D tracking.
The publicly available annotated RGB-D videos cannot support offline training of an RGB-D tracking network.
More specifically, while the existing datasets for RGB-only trackers contain thousands of video sequences with millions of annotated frames, the existing RGB-D datasets contain only 416 video sequences in total.

Recently, a new RGB-D tracking dataset named DepthTrack as well as an offline trained RGB-D tracker DeT have been made public~\cite{yan2021depthtrack}.
However, the training set of DepthTrack contains only 150 videos captured in realistic scenarios.
The vast majority of the RGB-D training data used for the development of tracker DeT was generated from RGB-only tracking datasets using monocular depth estimation techniques.
The real training data collected by the depth camera occupies only a very small proportion.
The performance of a deep RGB-D tracker trained in this way depends largely on the quality of the monocular depth estimation used for reconstructing the depth information.
In summary, the existing RGB-D data is far from sufficient to promote the rapid development of RGB-D tracking.

\begin{table*}[t]
\footnotesize
\centering
\caption{An overview of the existing RGB-D datasets. The ST/ LT means Short-Term/ Long-Term. }\label{table1}
\resizebox{6.8in}{!}{
\begin{tabular}{lcccccc}
\hline Dataset & Videos number & Total frames & Average length & Annotated frames & Scene attributes & ST/ LT   \\
\hline 
PTB & 100 & 21,542 & 215 & 21,542 &5& ST \\
STC & 36 & 9,009 & 250 & 9,009 &12 &ST  \\
CDTB & 80 & 101,956 & 1,274 & 101,956  & 13 &LT  \\
DepthTrack & 200 & 294,591& 1,473 & 294,591 & 15 &LT \\
RGBD1K & 1,050 & 2,503,400& 2,384 & 717,900 & 15 &LT \\
\hline
\end{tabular}}
\end{table*}

In order to further motivate the investigation of RGB-D tracking, and its use, we collect a new RGB-D dataset named RGBD1K.
RGBD1K contains 1,050 sequences with about 2.5M frames in total.  
Of these, 1,000 videos are reserved for training and 50 videos for testing. 
For the training videos, considering the annotation cost as well as the fact that the top one-fifth of frames of a long-term video contains representative visual and depth appearance variations for the learning of a tracking model, only the first 600 frames of each video are annotated.
Therefore, 600,000 annotated frames of RGBD1K can be utilized for supervised learning of deep RGB-D tracking methods.
Regarding the test set, all the frames are annotated, containing 117,900 frames in total.
Additionally, we annotate each frame with 15 challenging attributes.
The per-frame attributes facilitate the analysis of the trackers.
To the best of our knowledge, RGBD1K is the largest dataset for RGB-D tracking presently in existence.
The Table~\ref{table1} summarises the existing RGB-D datasets, including PTB~\cite{song2013tracking}, STC~\cite{xiao2017robust}, CDTB~\cite{lukezic2019cdtb}, DepthTrack~\cite{yan2021depthtrack} and our RGBD1K.
As evident from the table, the proposed RGBD1K has the largest number of videos, frames, annotations, and the average length of sequences.

In order to demonstrate the impact of the new dataset on the accuracy of RGB-D tracking, we propose a new baseline tracker based on spatial transformer learning, named SPT.
Specifically, we extend the RGB-only tracking network~\cite{yan2021learning} to an RGB-D version and introduce a novel fusion module designed to fuse the features from the two modalities. 
The SPT is trained offline using the 1,000 training videos of the RGBD1K dataset.
Extensive experiments, including the ablation experiments for self-analysis, and the comparative evaluation, are conducted on the RGBD1K, DepthTrack and CDTB datasets.
The corresponding results demonstrate the effectiveness of our RGBD1K dataset and the competitiveness of our new baseline tracker SPT.

\section{Related Work}
Recently, the advancement of RGB-D tracking has been stimulated by the emergence of RGB-D tracking datasets. 
In this section, we briefly introduce the techniques that are closely related to this tracking task.

\subsection{RGB-D tracking datasets}
There are four RGB-D object tracking datasets publicly available, including Princeton Tracking Benchmark (PTB)~\cite{song2013tracking}, Spatio-Temporal Consistency dataset (STC)~\cite{xiao2017robust}, Color and Depth Tracking Benchmark (CDTB)~\cite{lukezic2019cdtb} and DepthTrack~\cite{yan2021depthtrack}.
The specific properties of these four datasets are provided in Table~\ref{table1}.
PTB~\cite{song2013tracking} is the seminal publicly available RGB-D tracking dataset.
PTB contains 100 challenging videos recorded indoors for RGB-D tracking evaluation.
According to the target category, target size, movement, occlusion and motion type, these video sequences are labelled according to 11 attribute categories.
STC~\cite{xiao2017robust} is an RGB-D tracking dataset comprising 36 video sequences with 12 per-frame annotated attributes.
STC contains both indoor and outdoor scenarios.
CDTB~\cite{lukezic2019cdtb} is the existing largest test dataset for RGB-D tracking, comprising 80 video sequences captured in long-term tracking scenarios. 
All these sequences are annotated with 13 per-frame attributes.
CDTB dataset has been recently adopted in the VOT-RGBD 2019~\cite{kristan2019seventh}, 2020~\cite{kristan2020eighth} and 2021~\cite{kristan2021ninth} challenges.
DepthTrack~\cite{yan2021depthtrack} is the most recent RGB-D tracking dataset and is also the first dataset for offline training the RGB-D trackers.
It contains 200 RGB-D video sequences captured both indoors and outdoors, in which 150 videos can be adopted for offline learning. 
The remaining 50 videos are used for the tracking performance evaluation.

\begin{figure*}[!t]
\centering
\includegraphics[trim={0mm 64mm 0mm 0mm},clip,width=0.95\linewidth]{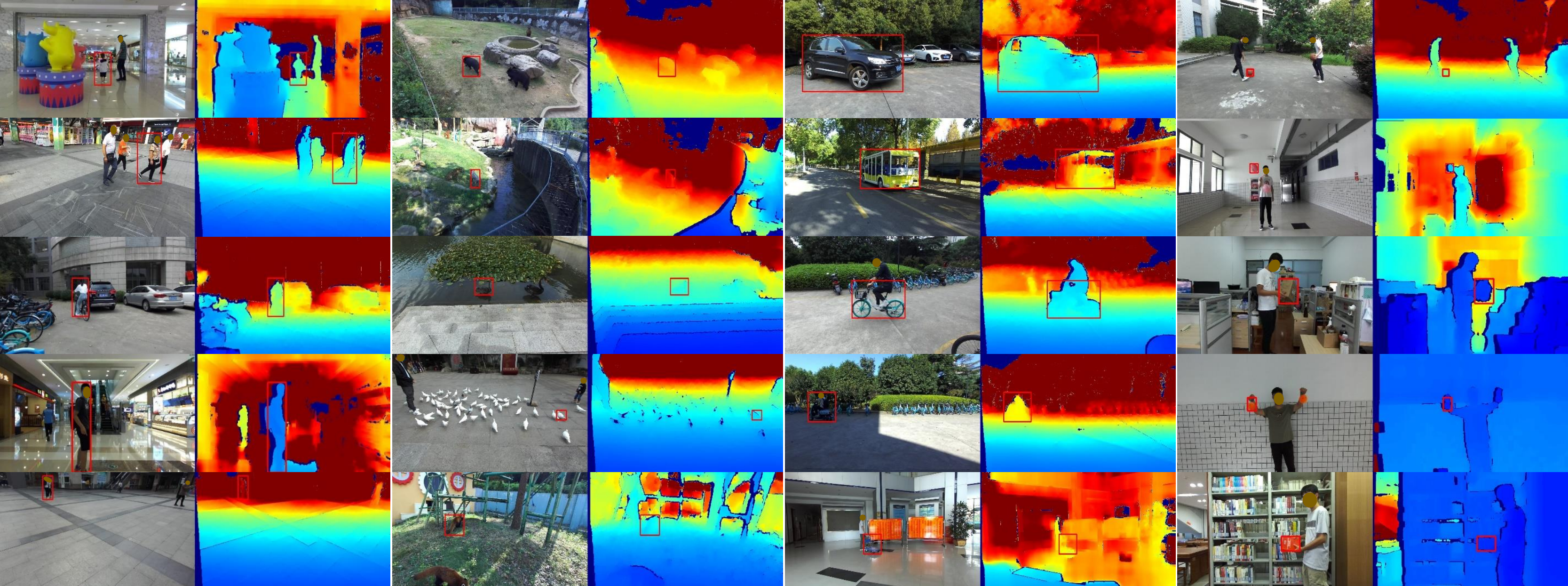}
\caption{RGB-D image samples from the RGBD1K. The targets are marked with red boxes, and the depth maps are converted to colour maps for more clear visualization. The first column is five samples of human, including \emph{child}, \emph{elder}, \emph{woman}, \emph{man} and \emph{couple}. The second column is five samples of animal, such as \emph{bear}, \emph{tiger}, \emph{swan}, \emph{pigeon} and \emph{red panda}. The third column is samples of vehicle, like \emph{car}, \emph{bus}, \emph{bicycle}, \emph{motorbike} and \emph{cart}. The final column is five samples of articles for daily use, including \emph{basketball}, \emph{balloon}, \emph{box}, \emph{doll} and \emph{book}.}
\label{object class}
\end{figure*}

\subsection{RGB-D tracking methods}
Since the advent of the seminal publicly available RGB-D tracking dataset PTB, the research on RGB-D tracking has received widespread attention.
Given the remarkable performance of the existing RGB-only tracking framework, most of the RGB-D tracking algorithms in the literature are simple RGB-only tracker extensions, utilizing the supplementary depth information effectively to improve the performance.

Based on the early colour-only Struck tracker and classical mean-shift tracker respectively, a local depth pattern feature~\cite{awwad2015local} and a 3-D mean-shift~\cite{liu2018context} are proposed for RGB-D tracking.
Besides, a 3D part-based sparse tracker with occlusion handling is developed from the particle filter framework~\cite{bibi20163d}.
Considering the promising performance of Discriminative Correlation Filter (DCF) based trackers on RGB videos~\cite{danelljan2017eco, xu2019joint, zhu2021robust}, the RGB-D trackers DS-KCF~\cite{camplani2015real}, DS-KCF-shape~\cite{hannuna2019ds}, DM-DCF~\cite{kart2018make} and OTR~\cite{kart2019object} are developed.
Recently, deep learning has been shown to exhibit promising performance in RGB-only tracking~\cite{bhat2019learning,  lukezic2020d3s, zhao2021adaptive, zhao2022distillation}.
The existing best performing RGB-D trackers are extensions of offline trained RGB-only trackers.
For example, in VOT-RGBD challenges~\cite{kristan2019seventh, kristan2020eighth, kristan2021ninth}, the trackers STARK\_RGBD, TALGD, ATCAIS are developed from the deep RGB trackers STARK~\cite{yan2021learning}, ATOM~\cite{danelljan2019atom} and DiMP~\cite{bhat2019learning}.
More recently, Yan~\textit{et al.} propose an end-to-end offline trained RGB-D tracker DeT~\cite{yan2021depthtrack}, which is based on the framework of the RGB-only trackers, ATOM and DiMP.

\begin{figure}[t]
\centering
\includegraphics[trim={15mm 45mm 15mm 52mm},clip,width=0.95\linewidth]{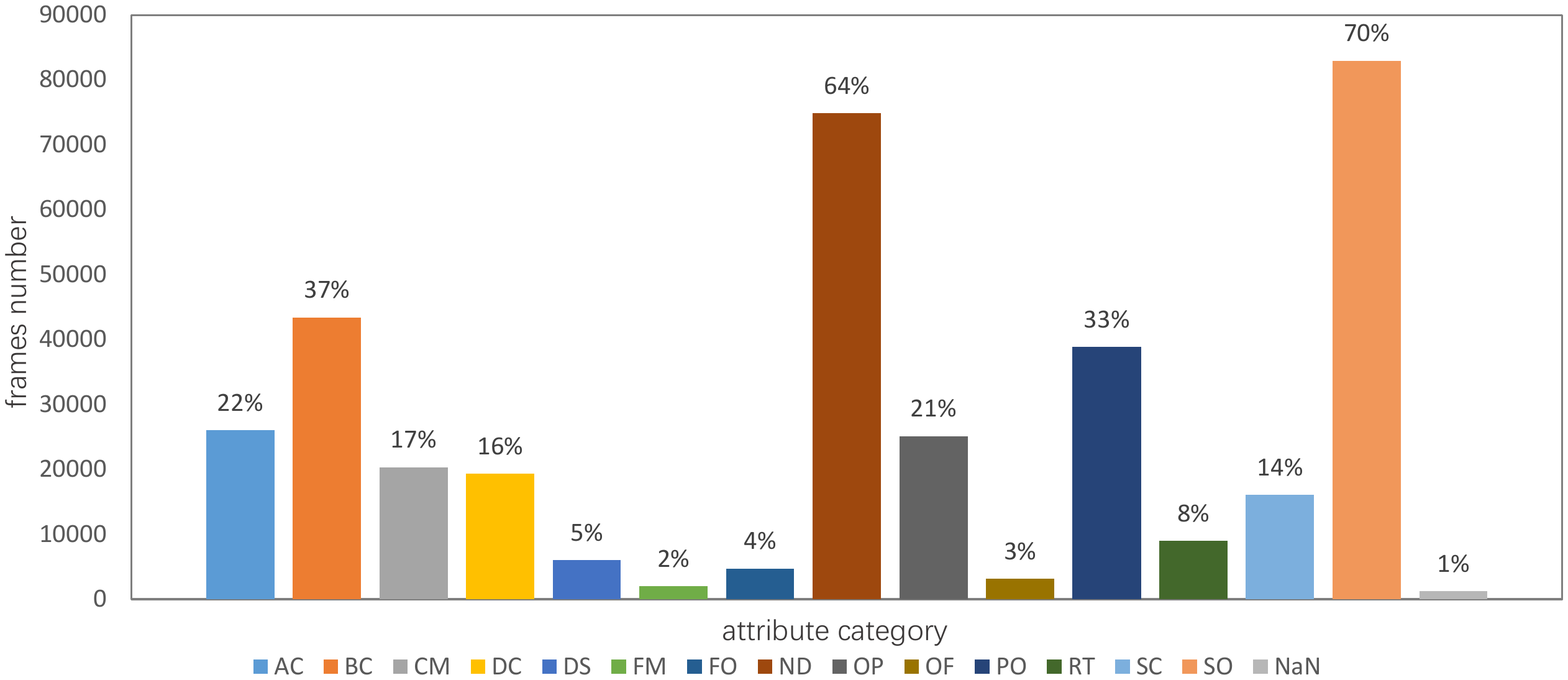}
\caption{The distribution of frames of different attributes in the RGBD1K test set.}
\label{attributes}
\end{figure}

\section{The RGBD1K Dataset}
\subsection{Video sequences}
RGBD1K contains 1,000 training sequences and 50 test sequences.
In total, the training set contains 2,385,500 frames and the test set contains 117,900 frames.
All the 1,050 sequences of our RGBD1K dataset are captured indoors and outdoors using the stereo camera ZED.
The ZED camera provides time-synchronized and pixel-aligned RGB and depth frames.
The video sequences share the same frame rate of 25 frames per second (fps).
The RGB images are stored using 24-bit (8-bit each channel) JEPG format, meanwhile, the depth maps are stored using 16-bit PNG format.

The RGBD1K covers a considerable number of object categories, including more than 100 different types concerned with humans, animals, vehicles and daily necessities.
Fig.~\ref{object class} provides some cases of different object classes.
We also select dozens of different scenes to record these sequences, such as office buildings, shopping malls, zoos, sports fields, etc.
Besides, some video sequences are captured from a first-person perspective and an overlooking perspective to simulate the perspectives of moving robots, UAVs and surveillance cameras.
For more statistical analysis and examples of different scenarios and object classes in the proposed RGBD1K dataset, please refer to the supplementary materials.

\begin{figure*}[t]
\centering
\includegraphics[trim={0mm 25mm 42mm 0mm},clip,width=0.95\linewidth]{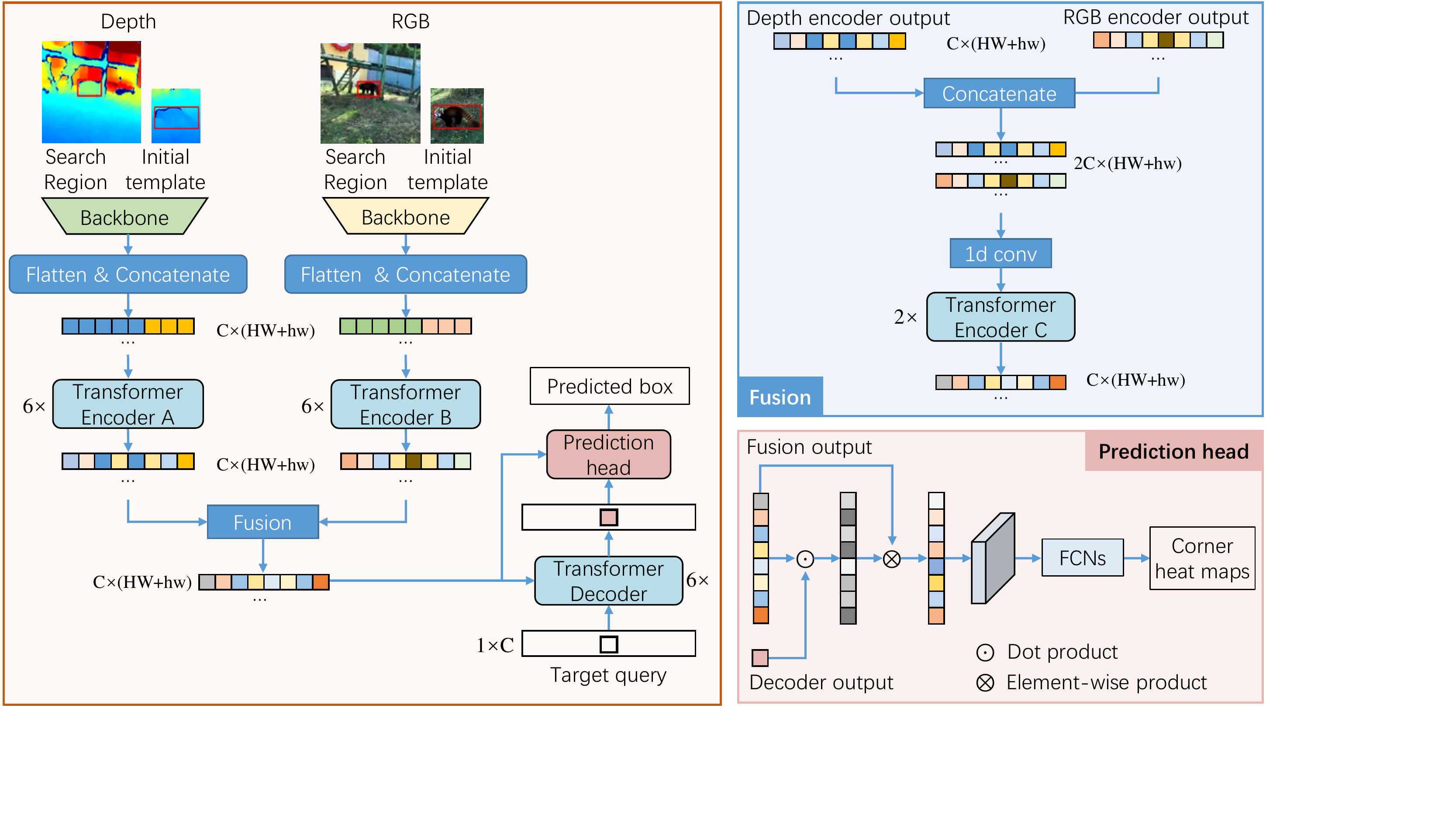}
\caption{Illustration of the framework of the proposed SPT tracker. The transformer encoder A and the transformer encoder B have the same structure, which stacks 6 encoder layers. The transformer encoder C stacks 2 encoder layers. 
}
\label{pipeline}
\end{figure*}

\subsection{Data annotation}
As to each video, we annotate the frames with the target bounding box.
It is universally acknowledged that data annotation is critical to research but time-consuming.
Considering that a short clip of a video sequence can contain sufficient visual and depth appearance variations, as well as to reduce the time cost, for the training set, we only annotate the frames of one segment of each video.
Specifically, we only annotate the first 600 frames of each sequence for the training set.
Although on average each video is only annotated with 1/4 of its length, we argue that the appearance variations in the annotated clips are sufficient for the learning of the spatio-temporal changing targets and scenarios \cite{valmadre2018long, kristan2018sixth}.
Besides, the unlabeled part is tightly related to the labelled part. 
Such partially annotated videos can be directly adopted for supervised learning, with the potential also to be utilized effectively for semi-supervised learning.
Meanwhile, it saves labour costs.
For the test set, all the frames of each sequence are annotated.

For further performance analysis of tracking methods, we annotate each frame of the test set with 15 attributes as proposed by CDTB~\cite{lukezic2019cdtb} and DepthTrack~\cite{yan2021depthtrack}, including Aspect-ratio Change (AC), Background Clutter (BC), Camera Motion (CM), Depth Change (DC), Dark Scene (DS), Fast Motion (FM), Full Occlusion (FO), Non-rigid Deformation (ND), Out-of-plane Rotation (OP), Out of Frame (OF), Partial Occlusion (PO), Reflective Target (RT), Size Change (SC), Similar Objects (SO) and Unassigned (NaN). 
The attributes AC, DC, FM, SC and NaN are calculated from the RGB-D images and the bounding box annotations. 
The remaining 10 attributes are annotated manually.
These scene attributes are beneficial for the trackers to analyse their merits and demerits in specific challenges.
For a detailed definition of each attribute, please refer to the supplementary materials.

The distribution of frames in each attribute category of the RGBD1K test set is reported in Fig.~\ref{attributes}.
From the figure, we can observe that only 1\% of the frames are marked without any scene attributes, which indicates that the RGBD1K test set is challenging.
Among the sequences, approximate 64\% of the frames are of non-rigid deformable targets. 
Typically, a deformable object implies a high probability of drastic appearance variations, which means it is more difficult for stable tracking.
In addition, 70\% frames are marked with the challenge attribute of similar objects.
The interference of similar objects in the background is an important issue worth studying for robust tracking.
Besides, background clutter and partial occlusion are also essential challenging factors in the RGBD1K test set.
Although some attributes contain a small number of frames, such as FO and OF only occupy 4\% and 3\% respectively, they are still very valuable for practical applications.
An RGB-D frame with the attribute of FO or OF means the target is invisible in the current frame.
Despite that the targets in only 7\% of the frames are invisible in total, this means that on average each video of the test set has about 165 frames of the target disappearance.
The frequent long period of target disappearance and reappearance complicates the tracking analysis, requiring perceptual capability for the RGBD tracker.

\begin{table*}[t]
\footnotesize
\centering
  \caption{A comparison of the STARK-S, STARK-S-FT and SPT on the RGBD1K, DepthTrack and CDTB datasets.}\label{ablation}
  \resizebox{6.5in}{!}{
    \begin{tabular}{l|ccc|ccc|ccc}
    \hline Dataset &  \multicolumn{3}{c|}{RGBD1K} &  \multicolumn{3}{c|}{DepthTrack} & \multicolumn{3}{c}{CDTB}    \\
    \hline
    \hline Method & Pr & Re & F-score & Pr & Re & F-score & Pr & Re & F-score  \\
    \hline 
    STARK-S    & 0.480 & 0.510 & 0.495   & 0.490 & 0.511 & 0.500   & 0.630 &  0.701  & 0.664\\
    STARK-S-FT & 0.509 & 0.537 & 0.522   & 0.497 & 0.517 & 0.507   & 0.638 &  0.706  & 0.670\\
    SPT        & 0.545 & 0.578 & 0.561   & 0.527 & 0.549 & 0.538   & 0.654 &  0.726  & 0.688 \\
    \hline
    \end{tabular}}
\end{table*}

\begin{table*}[t]
\footnotesize
\centering
\caption{The tracking results on the RGBD1K test set.}\label{comparison1}
\resizebox{6.5in}{!}{
\begin{tabular}{l|ccccccccc|c}
\hline Method & DDiMP & ATCAIS & DRefine & SLMD & DAL & DeT & TSDM & TALGD & Siam\_LTD & SPT \\
\hline 
Pr &      \bf{0.557} & 0.511 &  0.532 & 0.554 & 0.562 & 0.438 & 0.455 & 0.485 & 0.543 & 0.545\\
Re &      0.534 & 0.451 &  0.462 & 0.526 & 0.407 & 0.419 & 0.361 & 0.415 & 0.318 & \bf{0.578}\\
F-score & 0.545 & 0.479 &  0.494 & 0.540 & 0.472 & 0.428 & 0.403 & 0.447 & 0.398 & \bf{0.561}\\
ST/LT   & ST    & LT    &  LT    & LT&  LT   &  ST   & LT    &  LT   &  LT   &   ST\\
\hline
\end{tabular}}
\end{table*}

\begin{table*}[t]
\footnotesize
\centering
\caption{The tracking results on the DepthTrack dataset.}\label{comparison2}
\resizebox{6.5in}{!}{
\begin{tabular}{l|ccccccccc|c}
\hline Method & DDiMP & ATCAIS & CLGS\_D & SiamDW\_D & LTDSEd & Siam\_LTD & SiamM\_Ds & DAL & DeT & SPT \\
\hline 
Pr &      0.503 & 0.500 & 0.584 & 0.429 & 0.430 & 0.418 & 0.463 & 0.512 & \bf{0.560} & 0.527\\
Re &      0.469 & 0.455 & 0.369 & 0.436 & 0.382 & 0.342 & 0.264 & 0.369 & 0.506 & \bf{0.549}\\
F-score & 0.485 & 0.476 & 0.453 & 0.432 & 0.405 & 0.376 & 0.336 & 0.429 & 0.532 & \bf{0.538}\\
ST/LT   & ST    & LT    &  LT   & LT    &  LT   &  LT   & LT    & LT   &  ST   &   ST\\
\hline
\end{tabular}}
\end{table*}

\begin{table*}[t]
\footnotesize
\centering
\caption{The tracking results on the CDTB dataset.}\label{comparison3}
\resizebox{6.5in}{!}{
\begin{tabular}{l|ccccccccc|c}
\hline Method & DDiMP & ATCAIS & CLGS\_D & SiamDW\_D & LTDSEd & Siam\_LTD & SiamM\_Ds & OTR & DeT & SPT \\
\hline 
Pr &      0.703 &\bf{ 0.709} & 0.725 & 0.677 & 0.674 & 0.626 & 0.685 & 0.364 & 0.674 & 0.654\\
Re &      0.689 & 0.696 & 0.664 & 0.685 & 0.643 & 0.489 & 0.677 & 0.312 & 0.642 & \bf{0.726}\\
F-score & 0.696 & \bf{0.702} & 0.693 & 0.681 & 0.658 & 0.549 & 0.681 & 0.336 & 0.657 & 0.688\\
Speed (fps)&  4.7  & 1.3 & 7.3  & 3.8 & 5.7  &  13.0 & 19.4  &  1.8    &  36.8 &  25.3 \\
ST/LT   & ST    & LT    &  LT    & LT&  LT   &  LT   & LT    & LT   &  ST   &   ST\\
\hline
\end{tabular}}
\end{table*}

\subsection{Performance measures}
While there are no explicit restrictions on the use of RGBD1K, when evaluating trackers on the test set we advocate the use of the long-term tracking evaluation protocol from~\cite{lukevzivc2018now}, which is applied in the VOT-RGBD challenges~\cite{kristan2019seventh, kristan2020eighth, kristan2021ninth}.
The reason is that there are a certain proportion of frames in which the targets are invisible in the RGBD1K dataset, \textit{i.e.} the target may disappear and reappear several times in one video. 
For a tracker to be evaluated on RGBD1K, the ability to localise the target as well as to predict the target absence, and recapture the missed target, is of significance for a robust tracking system. 
Therefore, the long-term VOT evaluation protocol is precisely suited for evaluating trackers on our dataset.

The tracking Precision and Recall from~\cite{lukevzivc2018now} are applied as the performance measures.
Specifically, Precision is defined as the average overlap ratio of the predicted and ground truth targets on the frames where the target is detected.
The Recall represents the average overlap ratio of the predicted target bounding box and the ground truth annotation, measured on the frames where the target is visible.
The primary performance measure is F-score obtained by calculating the tracking F-measure that combines tracking Precision and Recall.
Besides, trackers can be conveniently evaluated on the RGBD1K by using the VOT challenge toolkit \cite{kristan2021ninth}.
For more details on the evaluation metrics, readers can refer to the literature~\cite{lukevzivc2018now} or our supplementary materials.

\section{A new Baseline RGB-D tracker}
To demonstrate the significance of the RGBD1K dataset as well as to inspire new designs for RGB-D tracking, we propose a new RGB-D tracking baseline coined as SPT.
The SPT is developed from the recent state-of-the-art transformer-based tracker STARK~\cite{yan2021learning}.
STARK is a distinguished RGB-only tracker, achieving remarkable performance on RGB-only tracking datasets.

The SPT is formed by extending the STARK-S (STARK without the temporal structure) to an RGB-D version with a dedicated feature fusion module.
The architecture of SPT is presented in Fig.~\ref{pipeline}.
Firstly, the search regions and the initial templates of the two modalities are input to the backbone to extract deep CNN features respectively. 
The backbone used here is the ResNet-50 network~\cite{he2016deep}.
The features of search regions and templates are of $H\times{W}\times{C}$ and $h\times{w}\times{C}$, respectively.
Then, the features of each modality are flattened and concatenated, following a 6-layer stacked transformer encoder to fuse the template-search appearance for the specific modality. 
Finally, the outputs of two modality-specific encoders are fused by our feature fusion module.

Here we introduce the proposed feature fusion module in detail.
Firstly, the depth encoder output and the RGB encoder output are concatenated across channels.
Then a $1d$ convolutional layer is adopted to reduce the channel number of the concatenated features from $2C$ to $C$. 
Finally, we introduce a transformer encoder stacking 2 encoder layers to further fuse and enhance the features of the two modalities.
Each encoder layer is composed of a multi-head self-attention module and a feed-forward network.

The rest parts of the framework include the target query, the transformer decoder and the target bounding box prediction head~\cite{yan2021learning}.
The transformer decoder, stacking 6 decoder layers, takes a learnable target query and the fused features as input.
Each decoder layer contains a self-attention, encoder-decoder attention, and a feed-forward network.
Later, the output of the transformer decoder and the fused features are fed into the bounding box prediction head to predict the target box coordinates.

In the bounding box prediction module, firstly, the decoder output is used to calculate the similarities with fused features, and the similarities are used to enhance the fused features.
Then the enhanced features are reshaped and passed through fully-convolutional networks to generate a top-left corner heat map and a bottom-right corner heat map.
With the top-left and bottom-right corner points, the object bounding box can be determined.
The loss function of SPT is the combination between $l_1$ loss and the IoU loss.
For more details on each component of SPT, please refer to the supplementary materials.

\section{Evaluation}
We perform extensive experiments on RGBD1K, DepthTrack and CDTB datasets.
In this section, we describe the implementation details of our tracker SPT, including the parameters setup and the experimental platform. 
Then, the results of ablation studies are presented, to demonstrate the effectiveness of our dataset as well as the proposed feature fusion module of the SPT tracker.
Finally, we provide the results and corresponding analysis of comparative experiments.

\subsection{Implementation details}
The proposed SPT tracker is trained and evaluated with an Intel i9-CPU and one NVIDIA GeForce RTX 3090 GPU.
The training and test parameters are set the same as Stark, except for the learning rate and training epoch number.
The learning rate is set as $10^{-5}$ and the total epoch number is 250.
As to the backbone, transformer encoder A, B, transformer decoder and box prediction head of SPT, we initialize their weights by using the weights of corresponding components of the officially published STARK-S model.
Then the SPT is trained on the training set of RGBD1K.
The tracking speed of SPT is about 25 fps.

\subsection{Ablation study}
In order to demonstrate the effectiveness of the proposed RGBD1K dataset for RGB-D tracking, firstly, we construct three trackers, including STARK-S~\cite{yan2021learning}, STARK-S-FT, and our SPT.
STARK-S, the STARK tracker with ResNet-50 as the backbone (without a temporal branch), is the baseline tracker of SPT.
We use the officially released trained model for STARK-S.
The STARK-S-FT is the STARK-S tracker fine-tuned on the RGBD1K using only all the RGB images of the training set.
The SPT is trained with the RGB-D images of RGBD1K.
The results of the test set of RGBD1K are provided in Table~\ref{ablation}.
Fine-tuned with the RGB images of the training set of RGBD1K, the STARK-S-FT improves the performance from 0.480, 0.510, and 0.495 to 0.509, 0.537 and 0.522 in terms of Precision, Recall and F-score, respectively.
Trained with RGB-D images of the training set of RGBD1K, SPT further improves the results to 0.545 for Precision, 0.578 for Recall and 0.561 for F-score.
This improvement enables us to draw the conclusion that the challenging RGB images, as well as the depth images of RGBD1K, are beneficial for improving RGB-D tracking performance.

To further confirm the merit of RGBD1K, we conduct the same experiments on two other datasets DepthTrack and CDTB, to explore the performance among STARK-S, STARK-S-FT and SPT.
It is worth noting that the trackers are trained only using the RGBD1K without sequences from DepthTrack or CDTB to fine-tune the tracking networks or corresponding hyper-parameters.
The results on DepthTrack and CDTB are exhibited in Table~\ref{ablation}.
After training on the RGBD1K dataset, the trackers STARK-S-FT and SPT achieve significant performance improvement on the DepthTrack and CDTB datasets.
Especially, the SPT trained with RGB-D data from the RGBD1K, improves the results of STARK-S from 0.490, 0.511 and 0.500 to 0.527, 0.549 and 0.538 and from 0.630, 0.701 and 0.664 to 0.654, 0.726 and 0.688 in terms of Precision, Recall and F-score on the DepthTrack and CDTB datasets, respectively.
Concerning the F-score measure, the SPT improves the STARK-S by \emph{7.6\%} and \emph{3.6\%} on DepthTrack and CDTB, respectively.
Apparently, the results shown in Table~\ref{ablation} demonstrate the generalised advantages of the proposed RGBD1K dataset in training end-to-end RGB-D trackers.
Furthermore, extensive ablation experiments and analyses about the fusion module are provided in the supplementary materials. 

\begin{figure*}[t]
\centering
\includegraphics[trim={0mm 0mm 0mm 0mm},clip,width=0.95\linewidth]{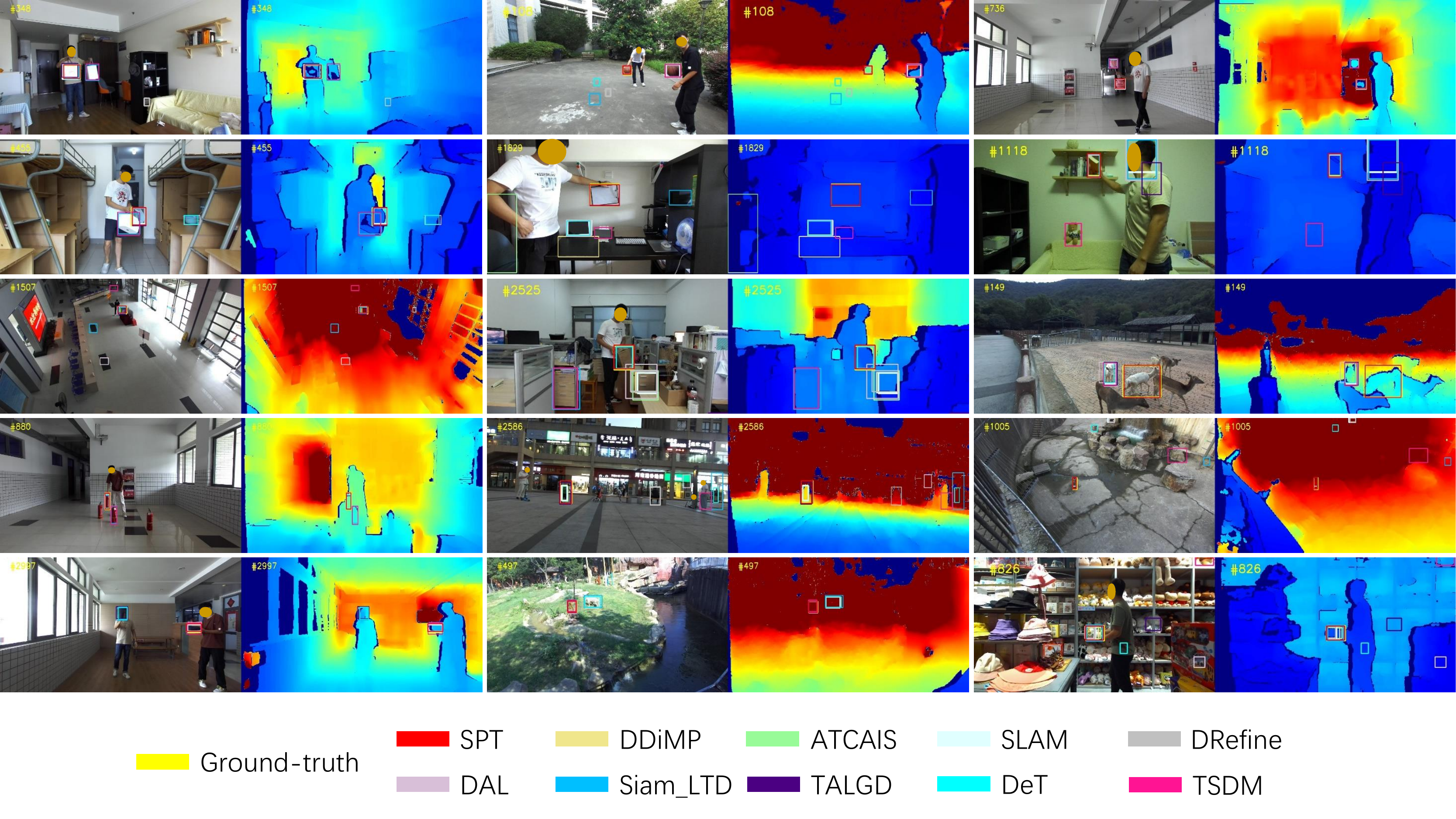}
\caption{An illustration of the qualitative experimental results on several challenging sequences of the \bb{RGBD1K} test set. Each RGB-D image is a sample from one particular sequence. The colour bounding boxes distinguish the ground-truth annotation and the results obtained by SPT, DDiMP, ATCAIS, SLAM, DRefine, DAL, Siam\_LTD, TALGD, DeT, and TSDM, respectively.}
\label{qualitative_comparison}
\end{figure*}

\subsection{Comparison with SOTA methods}
To demonstrate the superiority of the tracker SPT, we conduct quantitative, qualitative and attribute-based experiments with state-of-the-art trackers. 
The attribute-based experiments can be found in the supplementary materials.

\noindent{\bf{Quantitative Comparison:}}
We compare the proposed SPT with a considerable number of recent state-of-the-art RGB-D trackers on the RGBD1K test set.
In Table.~\ref{comparison1}, we report the results of RGB-D trackers, including DDiMP, ATCAIS and Siam\_LTD submitted to the VOT-RGBD 2020 challenge~\cite{kristan2020eighth}, TALGD, DRefine and SLMD from the VOT-RGBD 2021 challenge~\cite{kristan2021ninth}, DAL~\cite{qian2021dal}, DeT~\cite{yan2021depthtrack}, TSDM~\cite{zhao2021tsdm} and SPT.
Detailed results of some other RGB-only trackers on the RGBD1K test set are provided in the supplementary materials.
Generally, the F-score is the most important performance measure in the VOT protocol and the trackers are ranked according to F-score values.
As can be seen, on the RGBD1K test set, the SPT achieves the best F-score, and the short-term RGB-D tracker DDiMP is the second-best tracker.
Compared to the tracker DDiMP, our SPT tracker obtains \emph{2.9\%} improvement in terms of F-score.
Besides, compared with the long-term RGB-D trackers, such as ATCAIS, DRefine, SLMD and DAL, the proposed SPT tracker is also predominant, with gains of \emph{17.1\%}, \emph{13.5\%}, \emph{3.9\%}, and \emph{18.8\%} on F-score, respectively.
The tracking performance gain of the SPT indicates that training with the proposed RGBD1K facilitates more robust RGB-D tracking.

To further reflect the transferability and domain advantage of our RGBD1K, we compare the SPT tracker with state-of-the-art trackers on the DepthTrack and CDTB datasets.
It is still worth noting that our SPT tracker trained using RGBD1K is directly used to test on the DepthTrack and CDTB datasets without fine-tuning any parameters. 
The results on DepthTrack and CDTB are provided in Table.~\ref{comparison2} and \ref{comparison3}, respectively.

In Table.~\ref{comparison2}, the SPT tracker is compared with DDiMP, ATCAIS, CLGS\_D and Siam\_LTD from the VOT-RGBD 2020 challenge~\cite{kristan2020eighth}, SiamDW\_D, LTDSEd and  SiamM\_Ds from the VOT-RGBD 2019 challenge~\cite{kristan2019seventh}, DAL~\cite{qian2021dal} and DeT~\cite{yan2021depthtrack}.
From the results, the SPT achieves the best F-score of 0.538 and Recall of 0.549 on the DepthTrack dataset.
In Table.~\ref{comparison3}, the proposed SPT tracker is compared with DDiMP, ATCAIS, CLGS\_D SiamDW\_D, LTDSEd, Siam\_LTD, SiamM\_Ds, OTR~\cite{kart2019object} and DeT.
As can be seen, the SPT achieves significant superiority against the state-of-the-art trackers in terms of Recall.
Although the SPT achieves inferior Precision and F-score compared to DDiMP, ATCAIS and CLGS\_D, our SPT tracker has an obvious advantage in tracking speed.
The results provided above authenticate that the method offline trained with a large amount of real RGB-D data, such as the proposed baseline SPT tracker, can provide superior performance for RGB-D tracking. 
On the other hand, the results also confirm the significance of the proposed RGBD1K dataset for advanced RGB-D object tracking, although each video of RGBD1K is only annotated with the first 600 frames.

\noindent{\bf{Qualitative Comparison:}}
To intuitively display the advantages of our method, in Fig.~\ref{qualitative_comparison}, we provide a qualitative comparison of the tested RGB-D trackers, including SPT, DDiMP, ATCAIS, SLAM, DRefine, DAL, Siam\_LTD, TALGD, DeT, and TSDM, on several challenging videos from the RGBD1K dataset.
As can be seen in the figure, although suffering from different challenging factors, such as similar objects, partial occlusion, reappearing from full occlusion, camera motion, etc, our SPT can perform precise and steady tracking on these challenging videos.
Trained with additional depth images offline, the SPT can effectively alleviate the problem of similar objects, since two objects may vary in depth appearances when they are similar in visual appearance. 
Besides, the proposed fusion module in SPT enables to effectively fuse and enhance the features of RGB and depth modalities, making the depth information and the visual information complement each other, which helps the SPT to mitigate various complicated issues in RGB-D videos.
Therefore, undoubtedly, our SPT tracker can achieve promising RGB-D tracking performance.

\section{Conclusion}
In this work, we proposed a large-scale dataset for RGB-D tracking as well as a baseline tracker based on an end-to-end deep network.
This work is motivated by the scarcity of available annotated RGB-D videos that has hindered the development of RGB-D tracking. 
The proposed RGBD1K dataset contains more than twice as many videos as all the existing publicly available RGB-D videos for RGB-D object tracking.
To demonstrate the utility of the RGBD1K dataset, we designed a new baseline method named SPT for RGB-D tracking.
The SPT is trained offline using all the RGB-D videos of the training set of RGBD1K.
The extensive experimental results obtained using the RGBD1K test set, DepthTrack test set and CDTB dataset, have demonstrated the benefits of training on RGBD1K, and its capacity to promote the development of RGB-D trackers in the future.

\section{Acknowledgments}
This work was supported by the National Natural Science Foundation of China (62020106012, U1836218, 61672265), the 111 Project of Ministry of Education of China (B12018), and the Engineering and Physical Sciences Research Council (EPSRC) (EP/N007743/1, MURI/EPSRC/DSTL, EP/R018456/1).

\bibliography{rgbd1k}

\begin{thebibliography}{43}
\providecommand{\natexlab}[1]{#1}

\bibitem[{Awwad, Hussein, and Piccardi(2015)}]{awwad2015local}
Awwad, S.; Hussein, F.; and Piccardi, M. 2015.
\newblock Local depth patterns for tracking in depth videos.
\newblock In \emph{Proceedings of the 23rd ACM international conference on
  Multimedia}, 1115--1118.

\bibitem[{Bagautdinov, Fleuret, and Fua(2015)}]{bagautdinov2015probability}
Bagautdinov, T.; Fleuret, F.; and Fua, P. 2015.
\newblock Probability occupancy maps for occluded depth images.
\newblock In \emph{Proceedings of the IEEE Conference on Computer Vision and
  Pattern Recognition}, 2829--2837.

\bibitem[{Bhat et~al.(2019)Bhat, Danelljan, Gool, and
  Timofte}]{bhat2019learning}
Bhat, G.; Danelljan, M.; Gool, L.~V.; and Timofte, R. 2019.
\newblock Learning discriminative model prediction for tracking.
\newblock In \emph{Proceedings of the IEEE/CVF International Conference on
  Computer Vision}, 6182--6191.

\bibitem[{Bhat et~al.(2020)Bhat, Danelljan, Gool, and Timofte}]{bhat2020know}
Bhat, G.; Danelljan, M.; Gool, L.~V.; and Timofte, R. 2020.
\newblock Know your surroundings: Exploiting scene information for object
  tracking.
\newblock In \emph{European Conference on Computer Vision}, 205--221. Springer.

\bibitem[{Bibi, Zhang, and Ghanem(2016)}]{bibi20163d}
Bibi, A.; Zhang, T.; and Ghanem, B. 2016.
\newblock 3d part-based sparse tracker with automatic synchronization and
  registration.
\newblock In \emph{Proceedings of the IEEE Conference on Computer Vision and
  Pattern Recognition}, 1439--1448.

\bibitem[{Camplani et~al.(2015)Camplani, Hannuna, Mirmehdi, Damen, Paiement,
  Tao, and Burghardt}]{camplani2015real}
Camplani, M.; Hannuna, S.~L.; Mirmehdi, M.; Damen, D.; Paiement, A.; Tao, L.;
  and Burghardt, T. 2015.
\newblock Real-time RGB-D Tracking with Depth Scaling Kernelised Correlation
  Filters and Occlusion Handling.
\newblock In \emph{BMVC}, volume~3, 01--12.

\bibitem[{Chen et~al.(2021)Chen, Yan, Zhu, Wang, Yang, and
  Lu}]{chen2021transformer}
Chen, X.; Yan, B.; Zhu, J.; Wang, D.; Yang, X.; and Lu, H. 2021.
\newblock Transformer tracking.
\newblock In \emph{Proceedings of the IEEE/CVF Conference on Computer Vision
  and Pattern Recognition}, 8126--8135.

\bibitem[{Danelljan et~al.(2019)Danelljan, Bhat, Khan, and
  Felsberg}]{danelljan2019atom}
Danelljan, M.; Bhat, G.; Khan, F.~S.; and Felsberg, M. 2019.
\newblock Atom: Accurate tracking by overlap maximization.
\newblock In \emph{Proceedings of the IEEE/CVF Conference on Computer Vision
  and Pattern Recognition}, 4660--4669.

\bibitem[{Danelljan et~al.(2017)Danelljan, Bhat, Shahbaz~Khan, and
  Felsberg}]{danelljan2017eco}
Danelljan, M.; Bhat, G.; Shahbaz~Khan, F.; and Felsberg, M. 2017.
\newblock Eco: Efficient convolution operators for tracking.
\newblock In \emph{Proceedings of the IEEE conference on computer vision and
  pattern recognition}, 6638--6646.

\bibitem[{Danelljan, Gool, and Timofte(2020)}]{danelljan2020probabilistic}
Danelljan, M.; Gool, L.~V.; and Timofte, R. 2020.
\newblock Probabilistic regression for visual tracking.
\newblock In \emph{Proceedings of the IEEE/CVF conference on computer vision
  and pattern recognition}, 7183--7192.

\bibitem[{Fan et~al.(2019)Fan, Lin, Yang, Chu, Deng, Yu, Bai, Xu, Liao, and
  Ling}]{fan2019lasot}
Fan, H.; Lin, L.; Yang, F.; Chu, P.; Deng, G.; Yu, S.; Bai, H.; Xu, Y.; Liao,
  C.; and Ling, H. 2019.
\newblock Lasot: A high-quality benchmark for large-scale single object
  tracking.
\newblock In \emph{Proceedings of the IEEE conference on computer vision and
  pattern recognition}, 5374--5383.

\bibitem[{Griffiths et~al.(2017)Griffiths, Kalyanaraman, Ranjan, and
  Whitehouse}]{griffiths2017empirical}
Griffiths, E.; Kalyanaraman, A.; Ranjan, J.; and Whitehouse, K. 2017.
\newblock An empirical design space analysis of doorway tracking systems for
  real-world environments.
\newblock \emph{ACM Transactions on Sensor Networks (TOSN)}, 13(4): 1--34.

\bibitem[{Hannuna et~al.(2019)Hannuna, Camplani, Hall, Mirmehdi, Damen,
  Burghardt, Paiement, and Tao}]{hannuna2019ds}
Hannuna, S.; Camplani, M.; Hall, J.; Mirmehdi, M.; Damen, D.; Burghardt, T.;
  Paiement, A.; and Tao, L. 2019.
\newblock Ds-kcf: a real-time tracker for rgb-d data.
\newblock \emph{Journal of Real-Time Image Processing}, 16(5): 1439--1458.

\bibitem[{He et~al.(2016)He, Zhang, Ren, and Sun}]{he2016deep}
He, K.; Zhang, X.; Ren, S.; and Sun, J. 2016.
\newblock Deep residual learning for image recognition.
\newblock In \emph{Proceedings of the IEEE conference on computer vision and
  pattern recognition}, 770--778.

\bibitem[{Huang, Zhao, and Huang(2019)}]{huang2019got}
Huang, L.; Zhao, X.; and Huang, K. 2019.
\newblock Got-10k: A large high-diversity benchmark for generic object tracking
  in the wild.
\newblock \emph{IEEE Transactions on Pattern Analysis and Machine
  Intelligence}, 43(5): 1562--1577.

\bibitem[{Kart, Kamarainen, and Matas(2018)}]{kart2018make}
Kart, U.; Kamarainen, J.-K.; and Matas, J. 2018.
\newblock How to make an rgbd tracker?
\newblock In \emph{Proceedings of the European Conference on Computer Vision
  (ECCV) Workshops}, 01--15.

\bibitem[{Kart et~al.(2019)Kart, Lukezic, Kristan, Kamarainen, and
  Matas}]{kart2019object}
Kart, U.; Lukezic, A.; Kristan, M.; Kamarainen, J.-K.; and Matas, J. 2019.
\newblock Object tracking by reconstruction with view-specific discriminative
  correlation filters.
\newblock In \emph{Proceedings of the IEEE/CVF Conference on Computer Vision
  and Pattern Recognition}, 1339--1348.

\bibitem[{Kristan et~al.(2020)Kristan, Leonardis, Matas, Felsberg, Pflugfelder,
  K{\"a}m{\"a}r{\"a}inen, Danelljan, Zajc, Luke{\v{z}}i{\v{c}}, Drbohlav
  et~al.}]{kristan2020eighth}
Kristan, M.; Leonardis, A.; Matas, J.; Felsberg, M.; Pflugfelder, R.;
  K{\"a}m{\"a}r{\"a}inen, J.-K.; Danelljan, M.; Zajc, L.~{\v{C}}.;
  Luke{\v{z}}i{\v{c}}, A.; Drbohlav, O.; et~al. 2020.
\newblock The eighth visual object tracking VOT2020 challenge results.
\newblock In \emph{European Conference on Computer Vision}, 547--601. Springer.

\bibitem[{Kristan et~al.(2018)Kristan, Leonardis, Matas, Felsberg, Pflugfelder,
  ˇCehovin~Zajc, Vojir, Bhat, Lukezic, Eldesokey et~al.}]{kristan2018sixth}
Kristan, M.; Leonardis, A.; Matas, J.; Felsberg, M.; Pflugfelder, R.;
  ˇCehovin~Zajc, L.; Vojir, T.; Bhat, G.; Lukezic, A.; Eldesokey, A.; et~al.
  2018.
\newblock The sixth visual object tracking vot2018 challenge results.
\newblock In \emph{Proceedings of the European Conference on Computer Vision
  Workshops}, 01--52. Springer.

\bibitem[{Kristan et~al.(2019)Kristan, Matas, Leonardis, Felsberg, Pflugfelder,
  Kamarainen, Cehovin~Zajc, Drbohlav, Lukezic, Berg
  et~al.}]{kristan2019seventh}
Kristan, M.; Matas, J.; Leonardis, A.; Felsberg, M.; Pflugfelder, R.;
  Kamarainen, J.-K.; Cehovin~Zajc, L.; Drbohlav, O.; Lukezic, A.; Berg, A.;
  et~al. 2019.
\newblock The seventh visual object tracking vot2019 challenge results.
\newblock In \emph{Proceedings of the IEEE/CVF International Conference on
  Computer Vision Workshops}, 01--36.

\bibitem[{Kristan et~al.(2021)Kristan, Matas, Leonardis, Felsberg, Pflugfelder,
  K{\"a}m{\"a}r{\"a}inen, Chang, Danelljan, Cehovin, Luke{\v{z}}i{\v{c}}
  et~al.}]{kristan2021ninth}
Kristan, M.; Matas, J.; Leonardis, A.; Felsberg, M.; Pflugfelder, R.;
  K{\"a}m{\"a}r{\"a}inen, J.-K.; Chang, H.~J.; Danelljan, M.; Cehovin, L.;
  Luke{\v{z}}i{\v{c}}, A.; et~al. 2021.
\newblock The ninth visual object tracking vot2021 challenge results.
\newblock In \emph{Proceedings of the IEEE/CVF International Conference on
  Computer Vision}, 2711--2738.

\bibitem[{Li et~al.(2019)Li, Wu, Wang, Zhang, Xing, and Yan}]{li2019siamrpn++}
Li, B.; Wu, W.; Wang, Q.; Zhang, F.; Xing, J.; and Yan, J. 2019.
\newblock Siamrpn++: Evolution of siamese visual tracking with very deep
  networks.
\newblock In \emph{IEEE Conference on Computer Vision and Pattern Recognition},
  4282--4291.

\bibitem[{Liu et~al.(2018)Liu, Jing, Nie, Gao, Liu, and Jiang}]{liu2018context}
Liu, Y.; Jing, X.-Y.; Nie, J.; Gao, H.; Liu, J.; and Jiang, G.-P. 2018.
\newblock Context-aware three-dimensional mean-shift with occlusion handling
  for robust object tracking in RGB-D videos.
\newblock \emph{IEEE Transactions on Multimedia}, 21(3): 664--677.

\bibitem[{Lukezic et~al.(2019)Lukezic, Kart, Kapyla, Durmush, Kamarainen,
  Matas, and Kristan}]{lukezic2019cdtb}
Lukezic, A.; Kart, U.; Kapyla, J.; Durmush, A.; Kamarainen, J.-K.; Matas, J.;
  and Kristan, M. 2019.
\newblock Cdtb: A color and depth visual object tracking dataset and benchmark.
\newblock In \emph{Proceedings of the IEEE/CVF International Conference on
  Computer Vision}, 10013--10022.

\bibitem[{Lukezic, Matas, and Kristan(2020)}]{lukezic2020d3s}
Lukezic, A.; Matas, J.; and Kristan, M. 2020.
\newblock D3S-A discriminative single shot segmentation tracker.
\newblock In \emph{Proceedings of the IEEE/CVF Conference on Computer Vision
  and Pattern Recognition}, 7133--7142.

\bibitem[{Luke{\v{z}}i{\v{c}} et~al.(2018)Luke{\v{z}}i{\v{c}}, Zajc,
  Voj{\'\i}{\v{r}}, Matas, and Kristan}]{lukevzivc2018now}
Luke{\v{z}}i{\v{c}}, A.; Zajc, L.~{\v{C}}.; Voj{\'\i}{\v{r}}, T.; Matas, J.;
  and Kristan, M. 2018.
\newblock Now you see me: evaluating performance in long-term visual tracking.
\newblock \emph{arXiv preprint arXiv:1804.07056}.

\bibitem[{Mayer et~al.(2021)Mayer, Danelljan, Paudel, and
  Van~Gool}]{mayer2021learning}
Mayer, C.; Danelljan, M.; Paudel, D.~P.; and Van~Gool, L. 2021.
\newblock Learning target candidate association to keep track of what not to
  track.
\newblock In \emph{Proceedings of the IEEE/CVF International Conference on
  Computer Vision}, 13444--13454.

\bibitem[{Meshgi et~al.(2016)Meshgi, Maeda, Oba, Skibbe, Li, and
  Ishii}]{meshgi2016occlusion}
Meshgi, K.; Maeda, S.-i.; Oba, S.; Skibbe, H.; Li, Y.-z.; and Ishii, S. 2016.
\newblock An occlusion-aware particle filter tracker to handle complex and
  persistent occlusions.
\newblock \emph{Computer Vision and Image Understanding}, 150: 81--94.

\bibitem[{Muller et~al.(2018)Muller, Bibi, Giancola, Alsubaihi, and
  Ghanem}]{muller2018trackingnet}
Muller, M.; Bibi, A.; Giancola, S.; Alsubaihi, S.; and Ghanem, B. 2018.
\newblock Trackingnet: A large-scale dataset and benchmark for object tracking
  in the wild.
\newblock In \emph{Proceedings of the European Conference on Computer Vision
  (ECCV)}, 300--317.

\bibitem[{Qian et~al.(2021)Qian, Yan, Luke{\v{z}}i{\v{c}}, Kristan,
  K{\"a}m{\"a}r{\"a}inen, and Matas}]{qian2021dal}
Qian, Y.; Yan, S.; Luke{\v{z}}i{\v{c}}, A.; Kristan, M.;
  K{\"a}m{\"a}r{\"a}inen, J.-K.; and Matas, J. 2021.
\newblock DAL: A Deep Depth-Aware Long-term Tracker.
\newblock In \emph{2020 25th International Conference on Pattern Recognition
  (ICPR)}, 7825--7832. IEEE.

\bibitem[{Smeulders et~al.(2013)Smeulders, Chu, Cucchiara, Calderara, Dehghan,
  and Shah}]{smeulders2013visual}
Smeulders, A.~W.; Chu, D.~M.; Cucchiara, R.; Calderara, S.; Dehghan, A.; and
  Shah, M. 2013.
\newblock Visual tracking: An experimental survey.
\newblock \emph{IEEE transactions on pattern analysis and machine
  intelligence}, 36(7): 1442--1468.

\bibitem[{Song and Xiao(2013)}]{song2013tracking}
Song, S.; and Xiao, J. 2013.
\newblock Tracking revisited using RGBD camera: Unified benchmark and
  baselines.
\newblock In \emph{Proceedings of the IEEE international conference on computer
  vision}, 233--240.

\bibitem[{Valmadre et~al.(2018)Valmadre, Bertinetto, Henriques, Tao, Vedaldi,
  Smeulders, Torr, and Gavves}]{valmadre2018long}
Valmadre, J.; Bertinetto, L.; Henriques, J.~F.; Tao, R.; Vedaldi, A.;
  Smeulders, A.~W.; Torr, P.~H.; and Gavves, E. 2018.
\newblock Long-term tracking in the wild: A benchmark.
\newblock In \emph{Proceedings of the European conference on computer vision
  (ECCV)}, 670--685.

\bibitem[{Vaswani et~al.(2017)Vaswani, Shazeer, Parmar, Uszkoreit, Jones,
  Gomez, Kaiser, and Polosukhin}]{vaswani2017attention}
Vaswani, A.; Shazeer, N.; Parmar, N.; Uszkoreit, J.; Jones, L.; Gomez, A.~N.;
  Kaiser, {\L}.; and Polosukhin, I. 2017.
\newblock Attention is all you need.
\newblock \emph{Advances in neural information processing systems}, 30.

\bibitem[{Xiao et~al.(2017)Xiao, Stolkin, Gao, and Leonardis}]{xiao2017robust}
Xiao, J.; Stolkin, R.; Gao, Y.; and Leonardis, A. 2017.
\newblock Robust fusion of color and depth data for RGB-D target tracking using
  adaptive range-invariant depth models and spatio-temporal consistency
  constraints.
\newblock \emph{IEEE transactions on cybernetics}, 48(8): 2485--2499.

\bibitem[{Xu et~al.(2019)Xu, Feng, Wu, and Kittler}]{xu2019joint}
Xu, T.; Feng, Z.-H.; Wu, X.-J.; and Kittler, J. 2019.
\newblock Joint group feature selection and discriminative filter learning for
  robust visual object tracking.
\newblock In \emph{Proceedings of the IEEE/CVF International Conference on
  Computer Vision}, 7950--7960.

\bibitem[{Xue et~al.(2020)Xue, Li, Yin, Shang, Peng, and
  Shen}]{xue2020semantic}
Xue, X.; Li, Y.; Yin, X.; Shang, C.; Peng, T.; and Shen, Q. 2020.
\newblock Semantic-Aware Real-Time Correlation Tracking Framework for UAV
  Videos.
\newblock \emph{IEEE Transactions on Cybernetics}, 01--12.

\bibitem[{Yan et~al.(2021{\natexlab{a}})Yan, Peng, Fu, Wang, and
  Lu}]{yan2021learning}
Yan, B.; Peng, H.; Fu, J.; Wang, D.; and Lu, H. 2021{\natexlab{a}}.
\newblock Learning spatio-temporal transformer for visual tracking.
\newblock In \emph{Proceedings of the IEEE/CVF International Conference on
  Computer Vision}, 10448--10457.

\bibitem[{Yan et~al.(2021{\natexlab{b}})Yan, Yang, K{\"a}pyl{\"a}, Zheng,
  Leonardis, and K{\"a}m{\"a}r{\"a}inen}]{yan2021depthtrack}
Yan, S.; Yang, J.; K{\"a}pyl{\"a}, J.; Zheng, F.; Leonardis, A.; and
  K{\"a}m{\"a}r{\"a}inen, J.-K. 2021{\natexlab{b}}.
\newblock Depthtrack: Unveiling the power of rgbd tracking.
\newblock In \emph{Proceedings of the IEEE/CVF International Conference on
  Computer Vision}, 10725--10733.

\bibitem[{Zhao et~al.(2021{\natexlab{a}})Zhao, Liu, Wang, and
  Guo}]{zhao2021tsdm}
Zhao, P.; Liu, Q.; Wang, W.; and Guo, Q. 2021{\natexlab{a}}.
\newblock TSDM: Tracking by SiamRPN++ with a Depth-refiner and a
  Mask-generator.
\newblock In \emph{2020 25th International Conference on Pattern Recognition
  (ICPR)}, 670--676. IEEE.

\bibitem[{Zhao et~al.(2022)Zhao, Xu, Wu, and Kittler}]{zhao2022distillation}
Zhao, S.; Xu, T.; Wu, X.-J.; and Kittler, J. 2022.
\newblock Distillation, Ensemble and Selection for building a Better and Faster
  Siamese based Tracker.
\newblock \emph{IEEE Transactions on Circuits and Systems for Video
  Technology}, 01--13.

\bibitem[{Zhao et~al.(2021{\natexlab{b}})Zhao, Xu, Wu, and
  Zhu}]{zhao2021adaptive}
Zhao, S.; Xu, T.; Wu, X.-J.; and Zhu, X.-F. 2021{\natexlab{b}}.
\newblock Adaptive feature fusion for visual object tracking.
\newblock \emph{Pattern Recognition}, 111: 107679.

\bibitem[{Zhu et~al.(2021)Zhu, Wu, Xu, Feng, and Kittler}]{zhu2021robust}
Zhu, X.-F.; Wu, X.-J.; Xu, T.; Feng, Z.; and Kittler, J. 2021.
\newblock Robust Visual Object Tracking via Adaptive Attribute-Aware
  Discriminative Correlation Filters.
\newblock \emph{IEEE Transactions on Multimedia}, 24: 301--312.

\end{thebibliography}

\clearpage

\twocolumn[
\begin{@twocolumnfalse}
	\section*{\centering{Supplementary Material for \\ \emph{RGBD1K: A Large-scale Dataset and Benchmark for RGB-D Object Tracking}\\[35pt]}}
\end{@twocolumnfalse}
]

\section{Video Sequences}
\begin{figure*}[!t]
\centering
\includegraphics[trim={0mm 80mm 0mm 0mm},clip,width=0.95\linewidth]{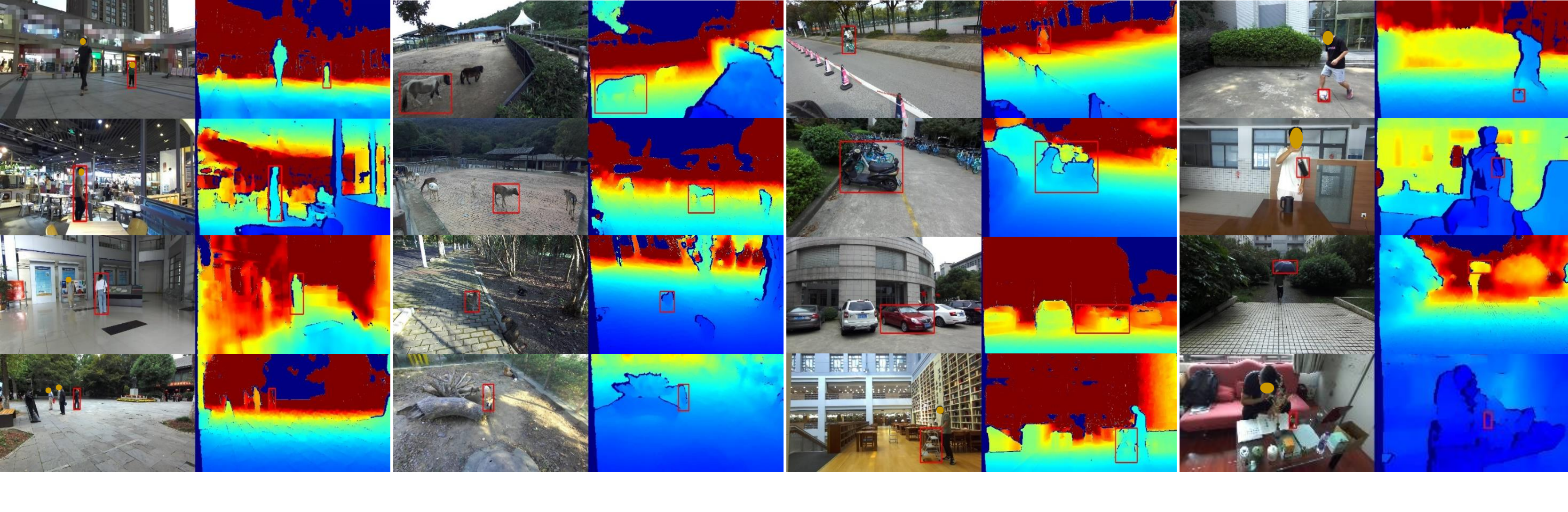}
\caption{RGB-D image samples from the RGBD1K. The targets are marked with red boxes, and the depth maps are converted to color maps for more clear visualization. The first column is four samples of human, including \emph{child}, \emph{man}, \emph{woman} and \emph{elder}. The second column is four samples of animal, such as \emph{horse}, \emph{deer}, \emph{cat}, and \emph{fox}. The third column is samples of vehicle, like \emph{bicycle}, \emph{motorbike}, \emph{car} and \emph{cart}. The final column is four samples of articles for daily use, including \emph{football}, \emph{cup}, \emph{umbrella}, and \emph{bottle}.}
\label{object class1}
\end{figure*}

\begin{figure*}[t]
\centering
\includegraphics[trim={0mm 80mm 0mm 0mm},clip,width=0.95\linewidth]{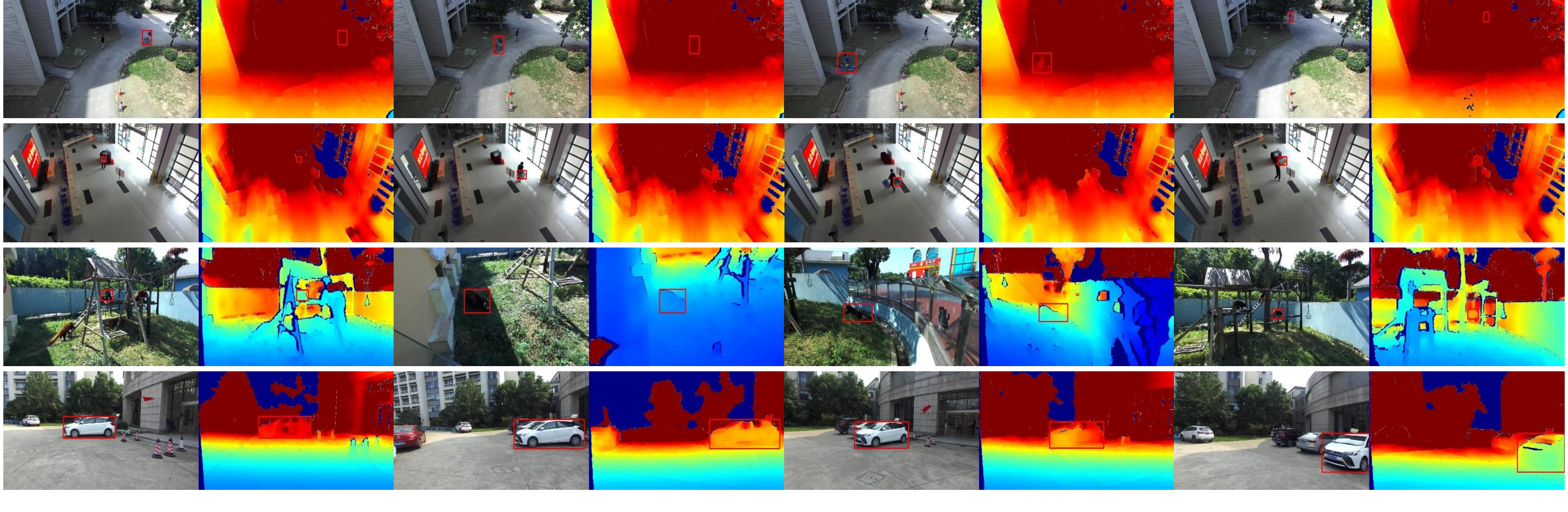}
\caption{RGB-D image samples from the RGBD1K. Each row is four RGB-D frames of one video. The first and second rows are videos recorded from overlooking perspective, while the third and fourth rows are videos recorded from first-person perspective.}
\label{perspectives}
\end{figure*}

To display the diversities of the objects and scenes of the proposed RGBD1K dataset, we provide more cases of various object classes and scene types in Fig.~\ref{object class1} and Fig.~\ref{perspectives}.
In Fig.~\ref{object class1}, some instances of different objects in the RGBD1K dataset are exhibited, involving the human, animal, vehicle, and articles for daily use.
Besides, these videos are recorded from various scenes, like plaza, canteen, park, office building, zoo, library, etc.
In Fig.~\ref{perspectives}, we present some video sequences captured from first-person perspective and overlooking perspective.
The first-person perspective and overlooking perspective can simulate the perspectives of moving robots, drones, and surveillance cameras, which makes the proposed RGBD1K more conducive to the research of real-world applications.
In addition, the camera motion and different perspectives of these videos lead to the more varied foreground and background appearance variations, making the proposed RGBD1K dataset more diverse and challenging.

\section{Visual Attributes}
A particular attribute denotes a specific challenging factor of a scenario.
Each frame of the RGBD1K test set is annotated with 15 attributes as proposed by CDTB~\cite{lukezic2019cdtb} and DepthTrack~\cite{yan2021depthtrack}.
The Aspect-ratio Change (AC), Depth Change (DC), Fast Motion (FM), Size Change (SC) are calculated by using the RGB-D images and corresponding target bounding boxes.

\begin{itemize}
    \item Aspect-ratio Change (AC): the ratio between the maximum and minimum aspect ratio (width/height) within 21 consecutive frames (10 frames before and after the current frame) is larger than 1.5.
    \item Depth Change (DC): the ratio between maximum and minimum of depth means of the target area within 21 consecutive frames (10 frames before and after the current frame) is larger than 1.5.
    \item Fast Motion (FM): the target center moves by at least 30\% of its size in consecutive frames.
    \item Size Change (SC): the ratio between the maximum and minimum target size in 21 consecutive frames (10 frames before and after the current frame) is larger than 1.5.
\end{itemize}
The other attributes, including Background Clutter (BC), Camera Motion (CM), Dark Scene (DS), Full Occlusion (FO), Non-rigid Deformation (ND), Out-of-plane Rotation (OP), Out of Frame (OF), Partial Occlusion (PO), Reflective Target (RT) and Similar Objects (SO) are assigned manually.
\begin{itemize}
    \item Background Clutter (BC): the target has a similar appearance as the nearby background. 
    \item Camera Motion (CM): the camera view is moving.
    \item Dark Scene (DS): the scenario is very dark.
    \item Full occlusion (FO): the target is fully occluded.
    \item Non-rigid Deformation (ND): the target can be deformed non-rigidly.
    \item Out-of-plane Rotation (OP): the target rotates out of the plane.
    \item Out of Frame (OF): the whole target leaves the view.
    \item Partial Occlusion (PO): the target is partially occluded.
    \item Reflective Target (RT): the surface of the target is reflective.
    \item Similar Objects (SO): there are objects similar to the target in the background.
\end{itemize}
Besides, frames not annotated with any of the aforementioned attributes are assigned with the attribute of Unassigned (NaN).

\section{Performance Measures}
We adopt a long-term tracking evaluation protocol from~\cite{lukevzivc2018now} to evaluate trackers on the RGBD1K dataset.
Tracking Precision (Pr) and Recall (Re) are implemented.

\begin{equation}\label{Pr}
Pr(\tau_\theta) = \frac{1}{N_p}\sum_{t\in\{t:A_t(\tau_\theta)\neq\emptyset\}}{\Omega(A_t(\tau_\theta), G_t)},\\
\end{equation}
\begin{equation}\label{Re}
Re(\tau_\theta) = \frac{1}{N_g}\sum_{t\in\{t:G_t\neq\emptyset\}}{\Omega(A_t(\tau_\theta), G_t)},
\end{equation}
where $G_t$ denotes the ground-truth bounding box and $A_t(\tau_\theta)$ denotes the predicted bounding box at frame $t$.
$\Omega(A_t(\tau_\theta), G_t)$ is the intersection-over-union (IoU) between the ground-truth and tracking prediction.
$\tau_\theta$ is a confidence threshold.
The evaluation protocol requires that a tracker reports bounding box predictions and confidence scores together.
If the predicted confidence score $\theta_t$ at frame $t$ is below $\tau_\theta$, then $A_t(\tau_\theta)=\emptyset$.
$N_p$ is the number of frames where the prediction is made, \textit{i.e.} $A_t(\tau_\theta)\neq\emptyset$, and $N_g$ is the number of frames where the target is visible, \textit{i.e.} $G_t\neq\emptyset$.
The tracking Precision and Recall are combined into F-score measure $F(\tau_\theta)$:
\begin{equation}\label{F-score}
F(\tau_\theta) = \frac{2Pr(\tau_\theta)Re(\tau_\theta)}{Pr(\tau_\theta)+Re(\tau_\theta)}.
\end{equation}

\begin{figure*}[t]
\centering
\includegraphics[trim={0mm 00mm 0mm 0mm},clip,width=0.95\linewidth]{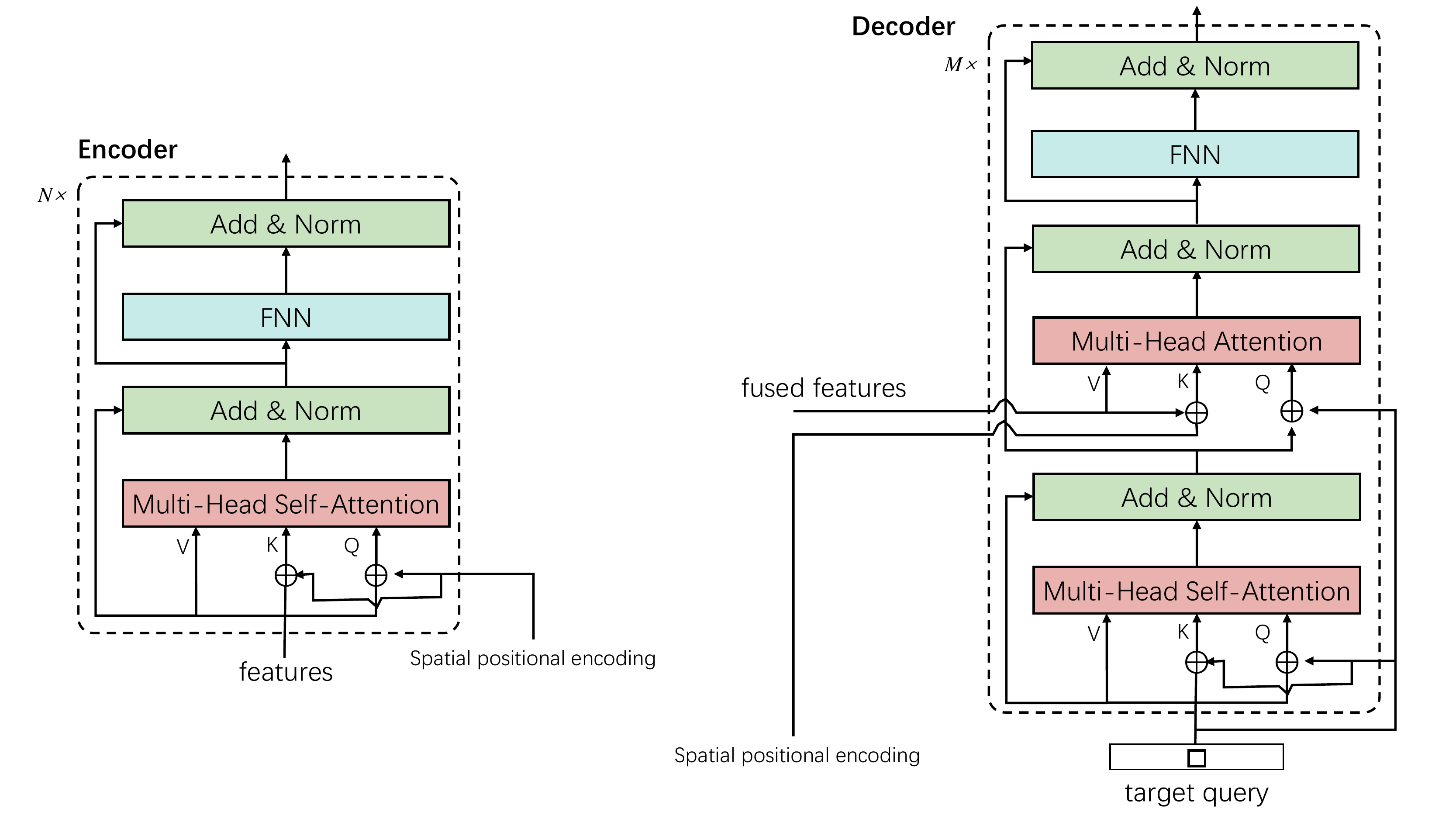}
\caption{Illustration of the architectures of transformer encoder and decoder.}
\label{encoder-decoder}
\end{figure*}

\section{RGB-D Tracking Baseline SPT}
The SPT is developed from the RGB-only tracker STARK~\cite{yan2021learning}.
The components, including the backbone, transformer encoder A, B, transformer decoder, bounding box prediction head, and the loss function are the same as the corresponding parts of STARK.\\
\textbf{Backbone.} The backbone of SPT adopts ResNet-50~\cite{he2016deep}.
The backbone network here has no other change with the original ResNet-50, except for removing the last stage and fully-connected layers.\\
\textbf{Encoder.} The transformer encoder A and encoder B have the same structure, which consists of 6 encoder layers.
Each encoder layer is made up of a multi-head self-attention module with a feed-forward network.
The transformer encoder C contains 2 encoder layers, each of which is composed of a multi-head self-attention module with a feed-forward network.
Besides, due to the permutation-invariance of the original transformer~\cite{vaswani2017attention}, sinusoidal positional embeddings are added to the input.
The structure of the transformer encoder is illustrated in Fig.~\ref{encoder-decoder}.\\
\textbf{Decoder.} The decoder takes a learnable target query and the fused features from the fusion module as the input. 
The transformer decoder stacks 6 decoder layers, each of which contains a self-attention, an encoder-decoder attention, and a feed-forward network. 
The structure of transformer decoder is illustrated in Fig.~\ref{encoder-decoder}.\\
\textbf{Prediction Head.} 
We first compute the similarities between the fused features and the output embedding from the decoder. 
Next, the similarity scores are element-wisely multiplied with the fused features to enhance important regions. 
Later, the enhanced features are reshaped to feature maps and then fed into a fully convolutional network (FCN). 
The FCN is composed of 5 stacked Conv-BN-ReLU layers and outputs two heat maps $P_{tl}(x, y)$ and $P_{br}(x, y)$ for the top-left and bottom-right corners of the object bounding box, respectively.
According to the two heat maps, the coordinates of top-left point $(\widehat{x_{tl}}, \widehat{y_{tl}})$ and bottom-right point $(\widehat{x_{br}}, \widehat{y_{br}})$ of the predicted box can be obtained by Eq.~\ref{corners}.

\begin{equation}\label{corners}
\begin{aligned}
(\widehat{x_{tl}}, \widehat{y_{tl}}) = (\sum_{y=0}^{H}\sum_{x=0}^{W}x\cdot{P_{tl}(x, y)}, \sum_{y=0}^{H}\sum_{x=0}^{W}y\cdot{P_{tl}(x, y))}, \\
(\widehat{x_{br}}, \widehat{y_{br}}) = (\sum_{y=0}^{H}\sum_{x=0}^{W}x\cdot{P_{br}(x, y)}, \sum_{y=0}^{H}\sum_{x=0}^{W}y\cdot{P_{br}(x, y))},
\end{aligned}
\end{equation}
where $W$ and $H$ are the width and height of the heat maps. With the top-left and bottom-right corners, the target bounding box can be determined. \\
\textbf{Loss Function.} 
The loss function of SPT is the combination of the $l_1$ loss and the IoU loss, as
\begin{equation}\label{loss}
L = \lambda_{iou}L_{iou}(b_i, \hat{b}_i) + \lambda_{L_1}L_1(b_i, \hat{b}_i),
\end{equation}
where $b_i$ and $\hat{b}_i$ are the ground-truth and the predicted box respectively, and $\lambda_{iou}$ and $\lambda_{L_1}$ are two constants.

\begin{figure*}[t]
\centering
\includegraphics[trim={0mm 82mm 0mm 0mm},clip,width=0.98\linewidth]{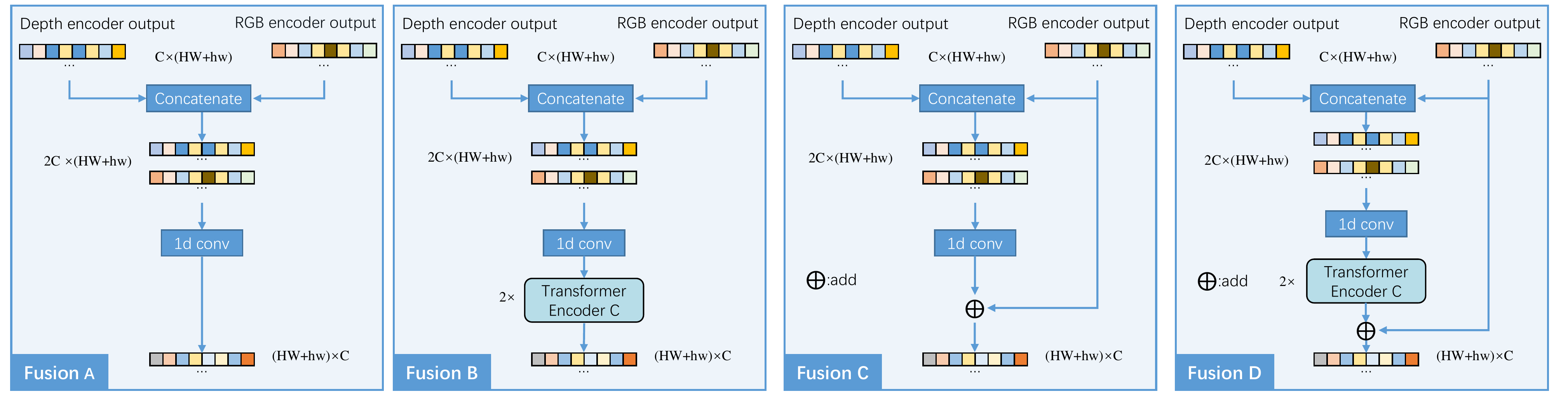}
\caption{Illustration of four different structures of feature fusion module. In the proposed SPT tracker, the second structure Fusion B is adopted.}
\label{fusion_modules}
\end{figure*}

\begin{table}[t]
    \footnotesize
    \centering
      \caption{Experimental results of four different fusion structures on RGBD1K.}\label{fuse_rgbd1k}
        \begin{tabular}{l|ccc}
        \hline Method & Pr & Re & F-score   \\
        \hline 
        Fusion A & 0.515 & 0.545 & 0.530  \\
        Fusion B & 0.545 & 0.578 & 0.561  \\
        Fusion C & 0.516 & 0.546 & 0.531    \\
        Fusion D & 0.548 & 0.580 & 0.563    \\
        \hline
        \end{tabular}
 \end{table}
 
\begin{table}[t]
  \footnotesize
  \centering
      \caption{Experimental results of four different fusion structures on CDTB.}\label{fuse_cdtb}
        \begin{tabular}{l|ccc}
        \hline Method & Pr & Re & F-score   \\
        \hline 
        Fusion A & 0.642 & 0.711 & 0.675  \\
        Fusion B & 0.654 & 0.726 & 0.688  \\
        Fusion C & 0.647 & 0.716 & 0.680    \\
        Fusion D & 0.647 & 0.717 & 0.681    \\
        \hline
        \end{tabular}
\end{table}

\begin{table*}[t]
\footnotesize
\centering
\caption{The tracking results on the RGBD1K test set.}\label{comparison}
\resizebox{6.5in}{!}{
\begin{tabular}{l|ccccccccc}
\hline Method & STARK & PrDiMP & DiMP & SiamRPN++ & KeepTrack & KYS & TransT & D3S & SPT \\
\hline 
Pr &      0.481 & 0.393 & 0.408 & 0.392 & 0.509 & 0.375 & \textbf{0.581} & 0.355 & 0.545 \\
Re &      0.509 & 0.415 & 0.430 & 0.411 & 0.541 & 0.394 & 0.443 & 0.342 & \textbf{0.578} \\
F-score & 0.495 & 0.404 & 0.419 & 0.401 & 0.525 & 0.384 & 0.502 & 0.348 & \textbf{0.561} \\
\hline
\end{tabular}}
\end{table*}

\begin{figure*}[t]
\centering
\includegraphics[trim={0mm 18mm 0mm 0mm},clip,width=0.95\linewidth]{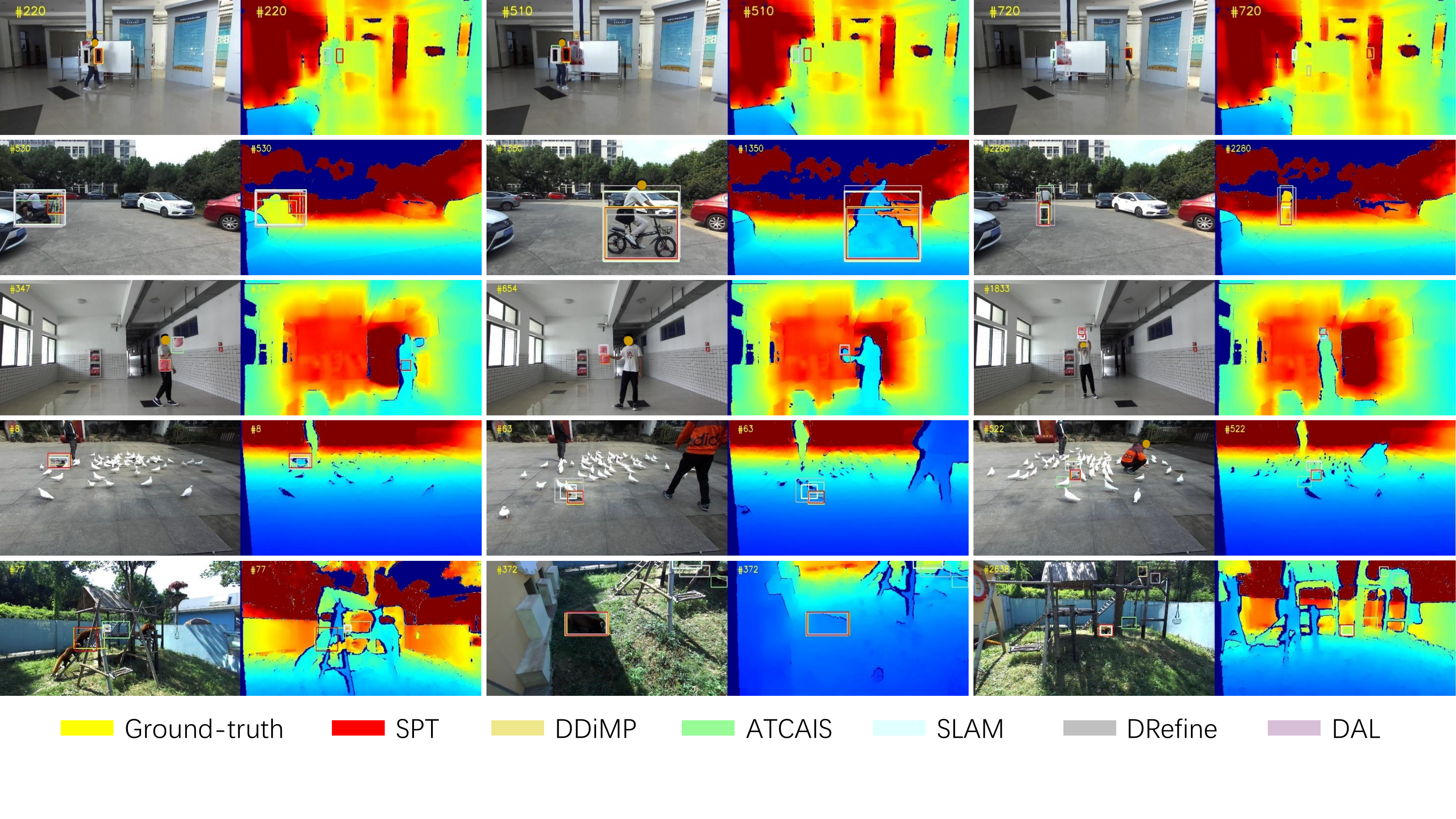}
\caption{An illustration of the qualitative experimental results on several challenging sequences of the RGBD1K test set. The colour bounding boxes distinguish the ground-truth annotation and the results obtained by SPT, DDiMP, ATCAIS, SLAM, DRefine and DAL, respectively.}
\label{qualitative_comparison1}
\end{figure*}

\begin{figure*}[t]
\centering
\includegraphics[trim={46mm 26mm 52mm 30mm},clip,width=0.24\linewidth]{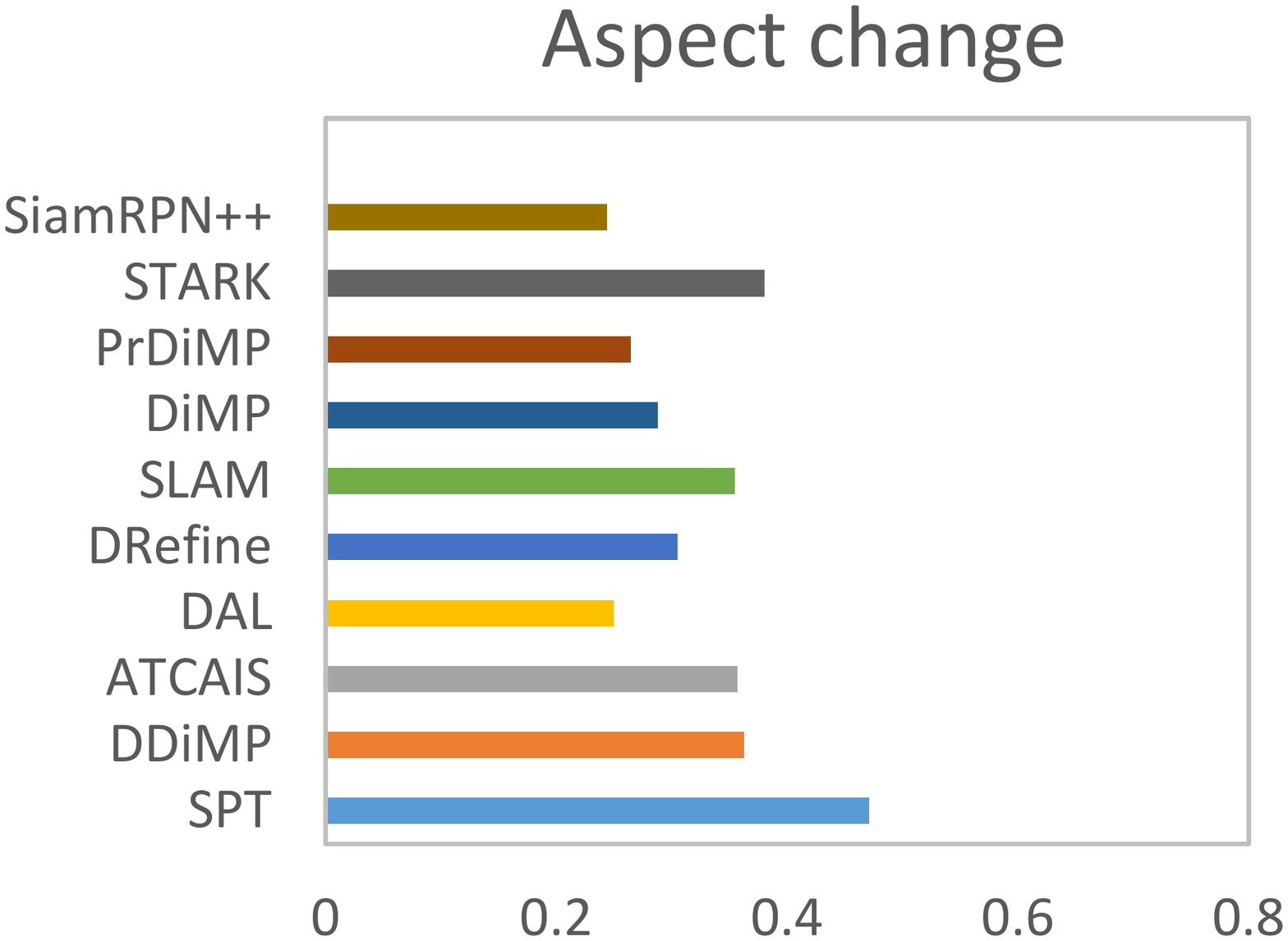}
\includegraphics[trim={46mm 26mm 52mm 30mm},clip,width=0.24\linewidth]{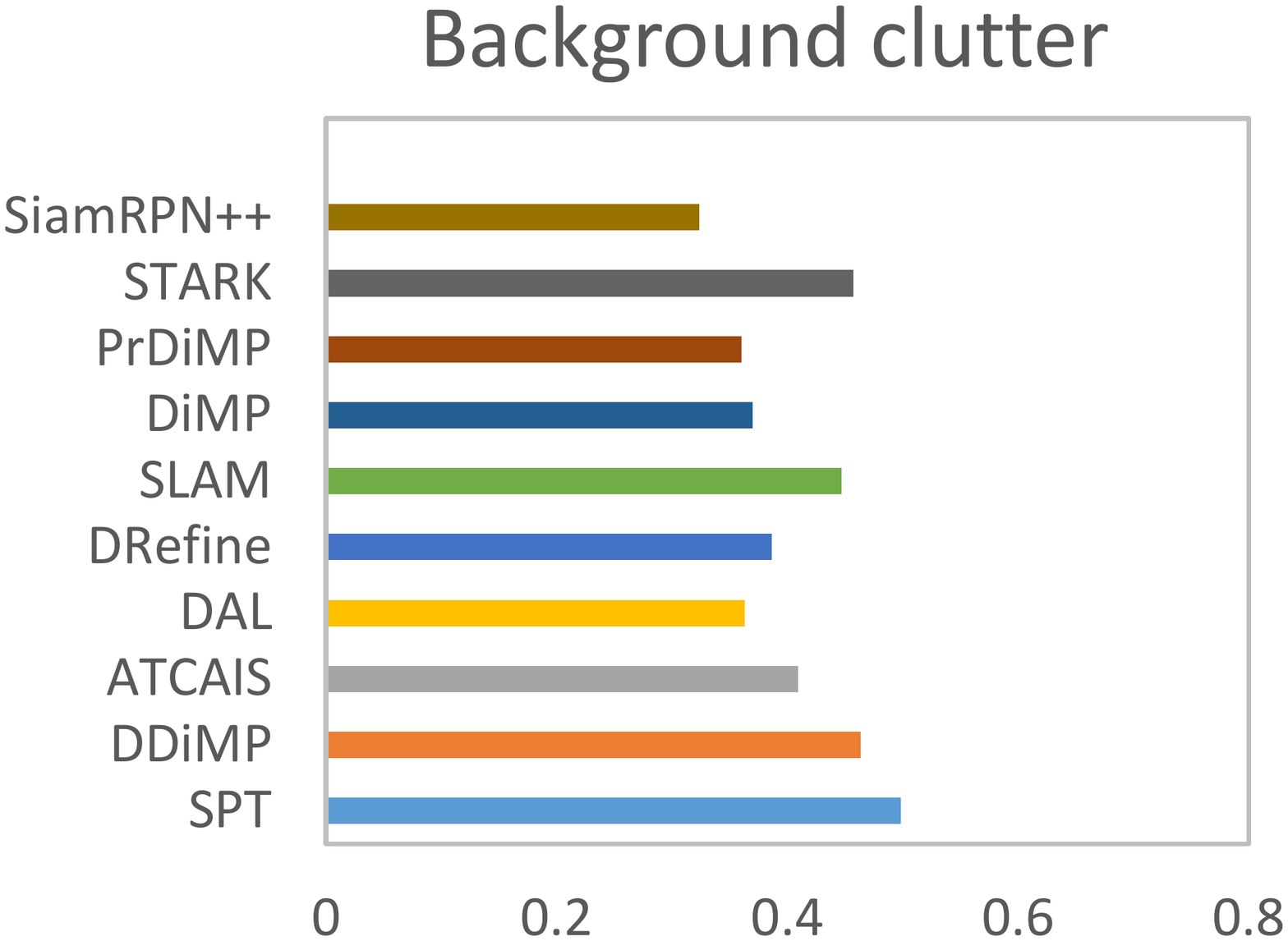}
\includegraphics[trim={46mm 26mm 52mm 30mm},clip,width=0.24\linewidth]{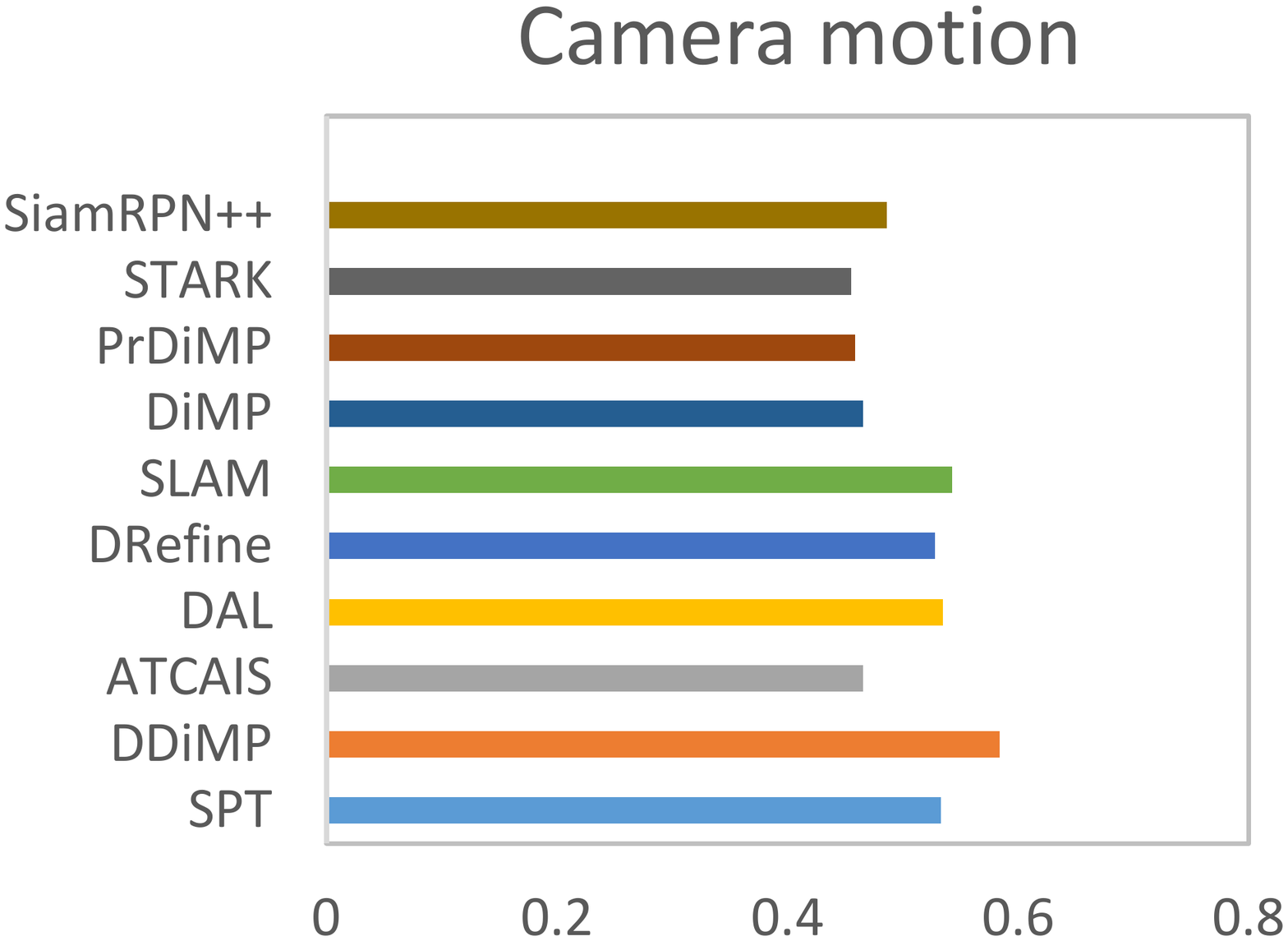}
\includegraphics[trim={46mm 26mm 52mm 30mm},clip,width=0.24\linewidth]{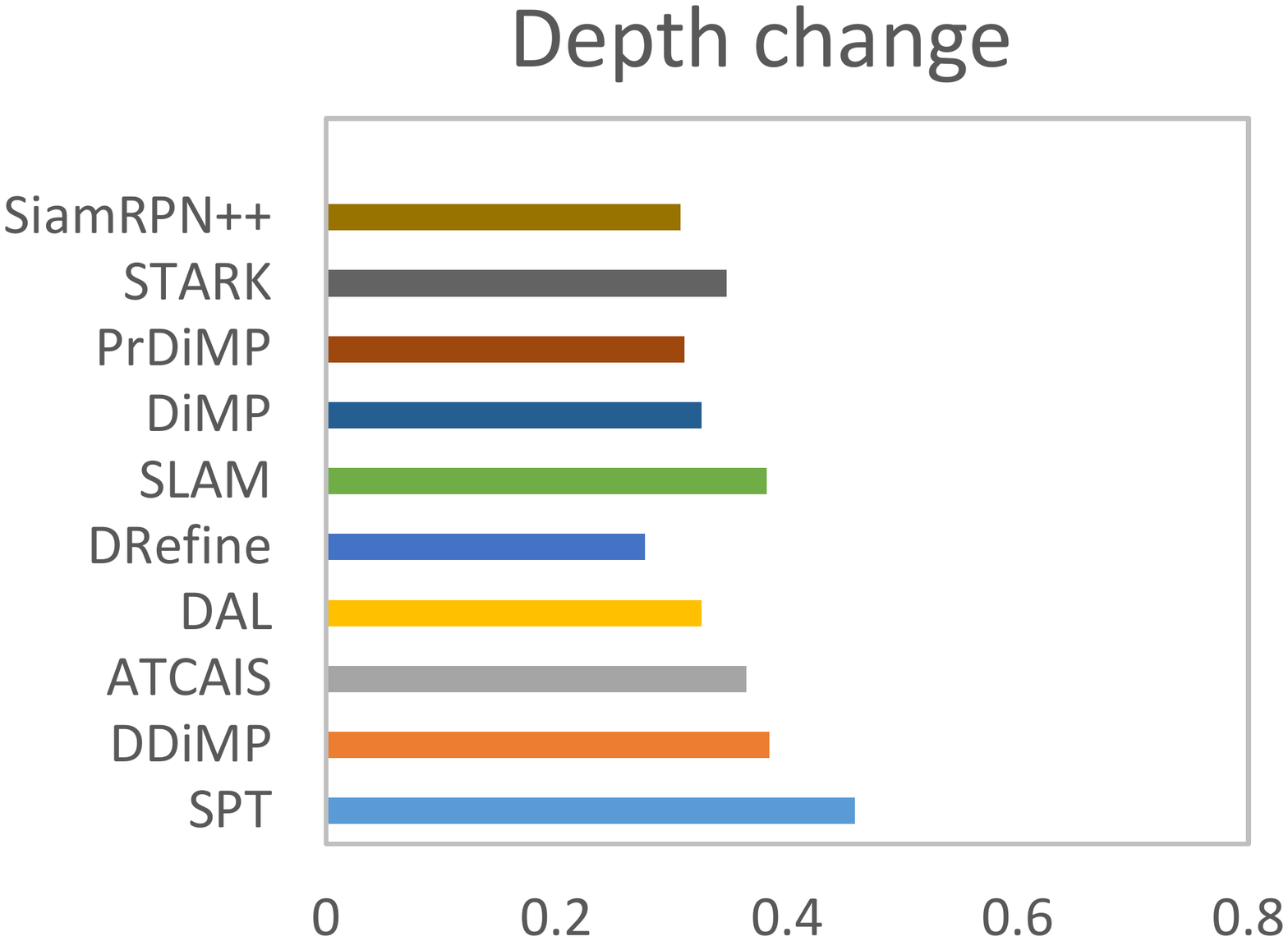}\\
\includegraphics[trim={46mm 26mm 52mm 30mm},clip,width=0.24\linewidth]{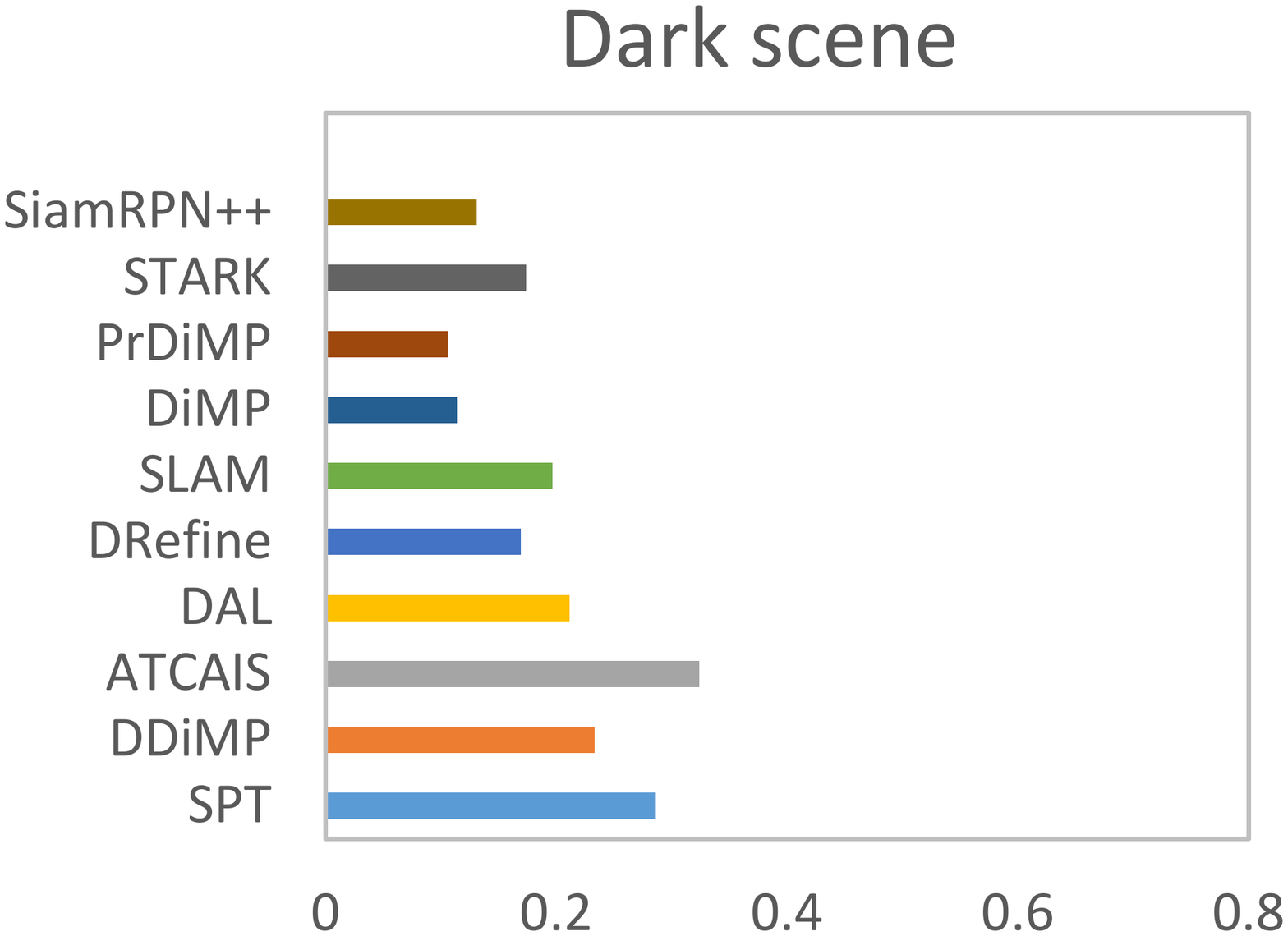}
\includegraphics[trim={46mm 26mm 52mm 30mm},clip,width=0.24\linewidth]{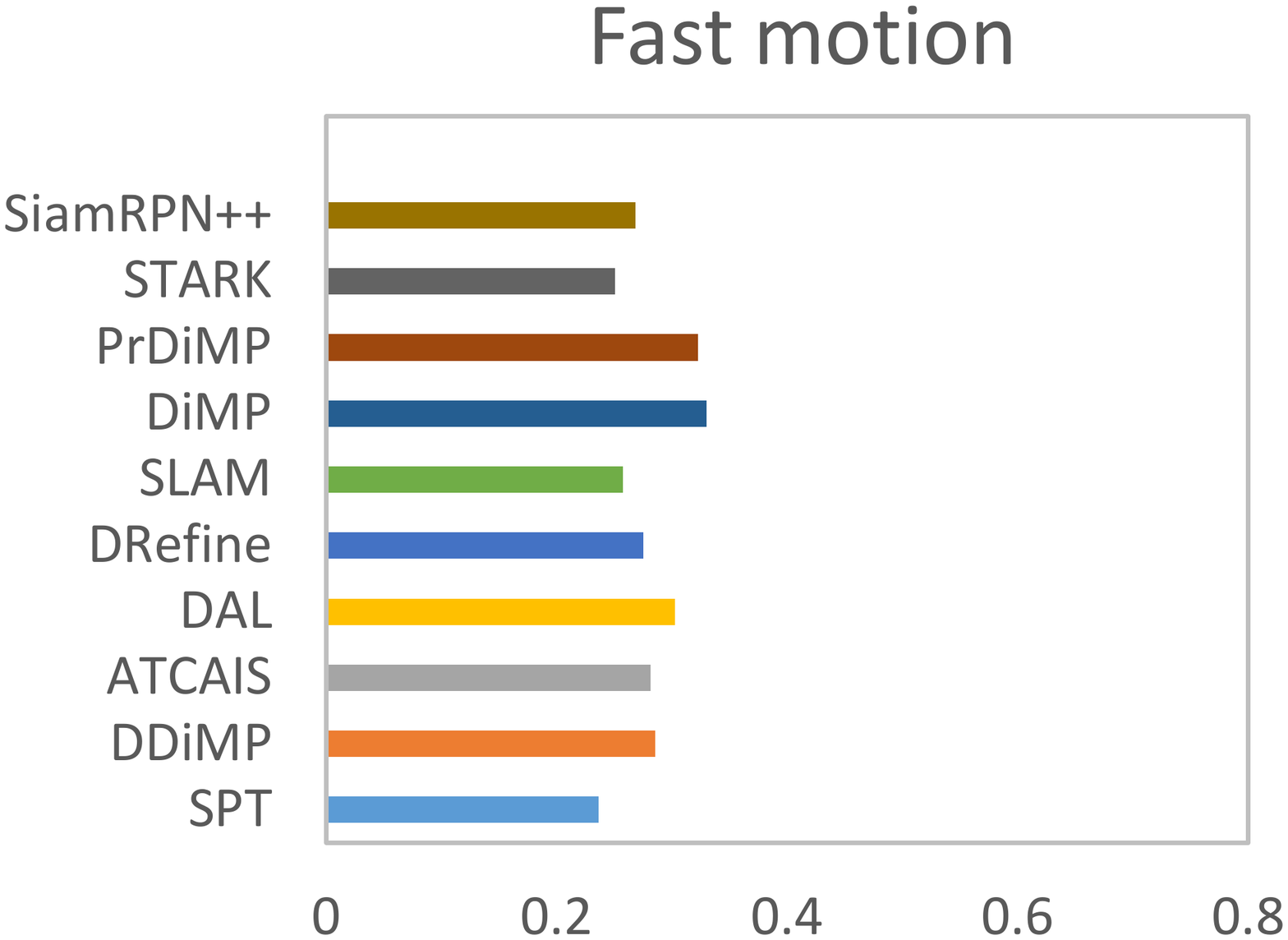}
\includegraphics[trim={46mm 26mm 52mm 30mm},clip,width=0.24\linewidth]{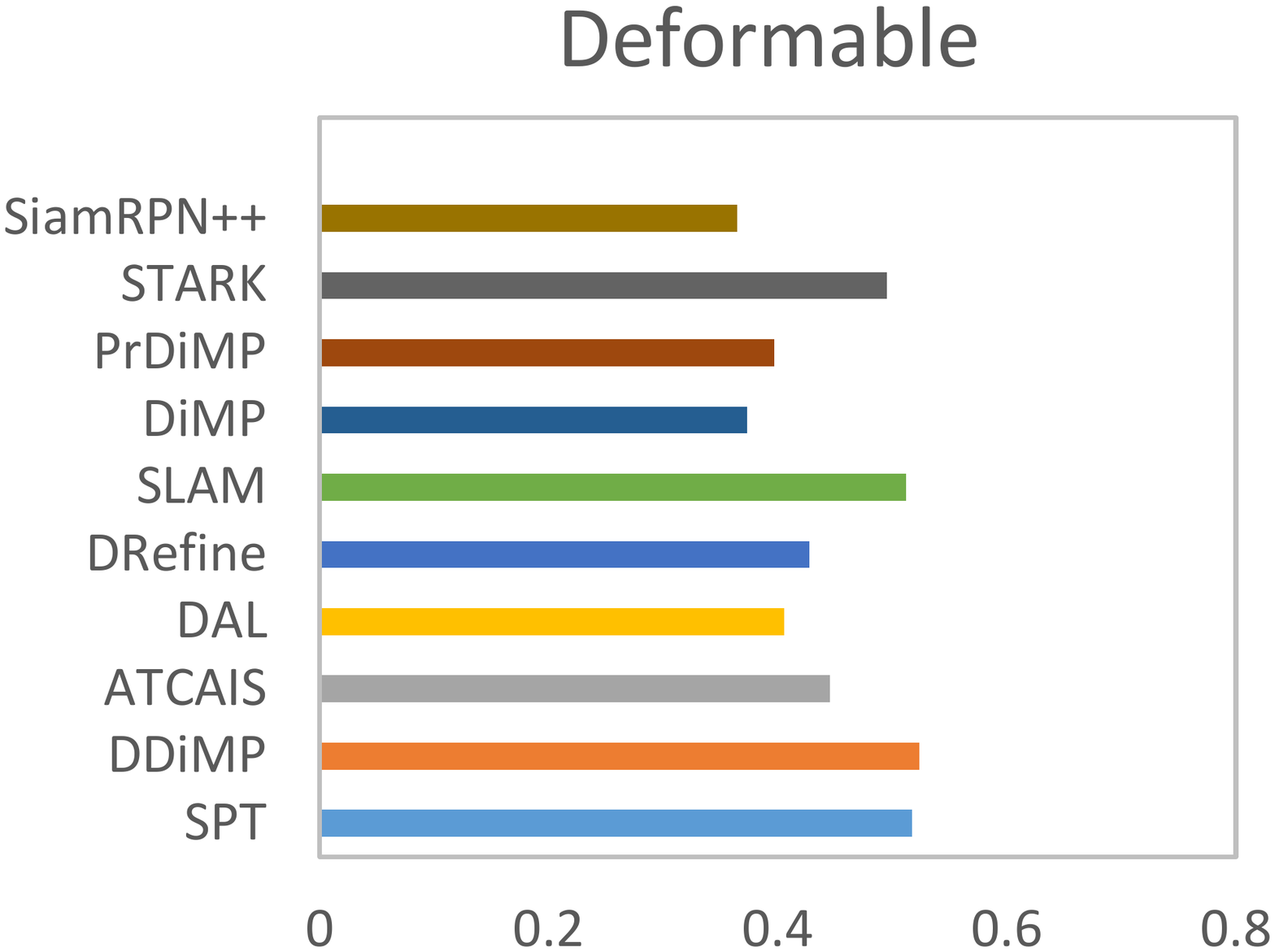}
\includegraphics[trim={46mm 26mm 52mm 30mm},clip,width=0.24\linewidth]{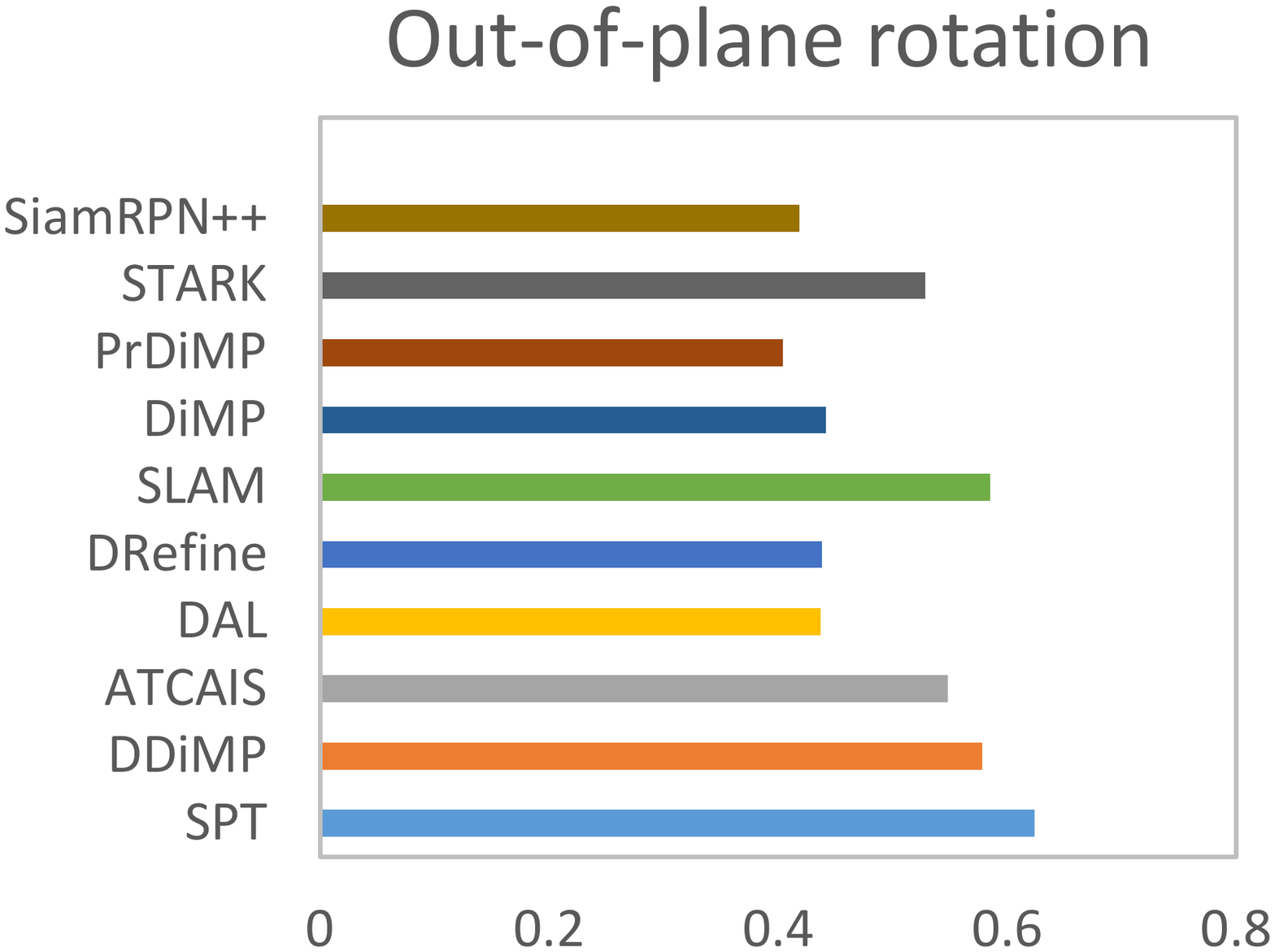}\\
\includegraphics[trim={46mm 26mm 52mm 30mm},clip,width=0.24\linewidth]{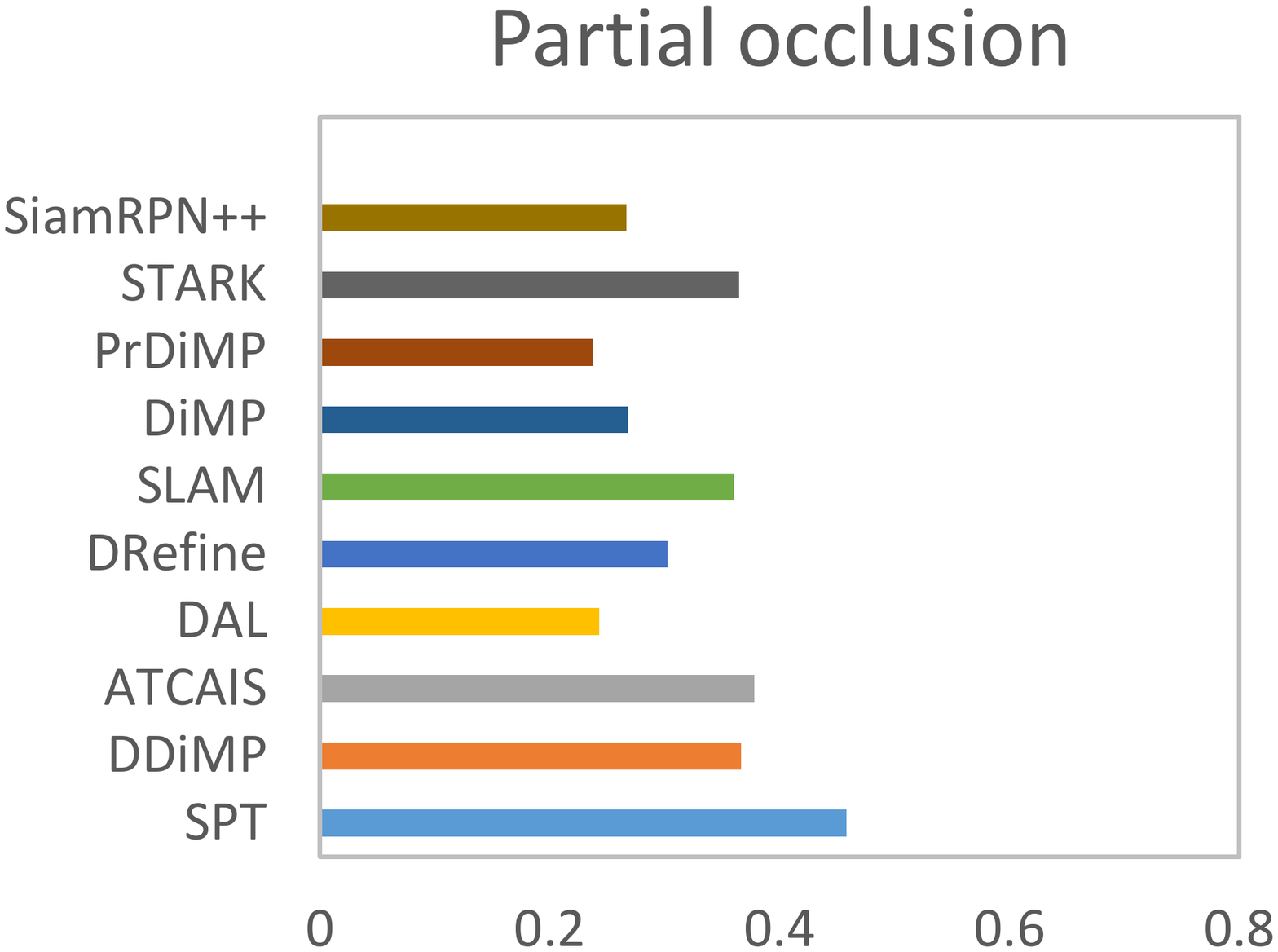}
\includegraphics[trim={46mm 26mm 52mm 30mm},clip,width=0.24\linewidth]{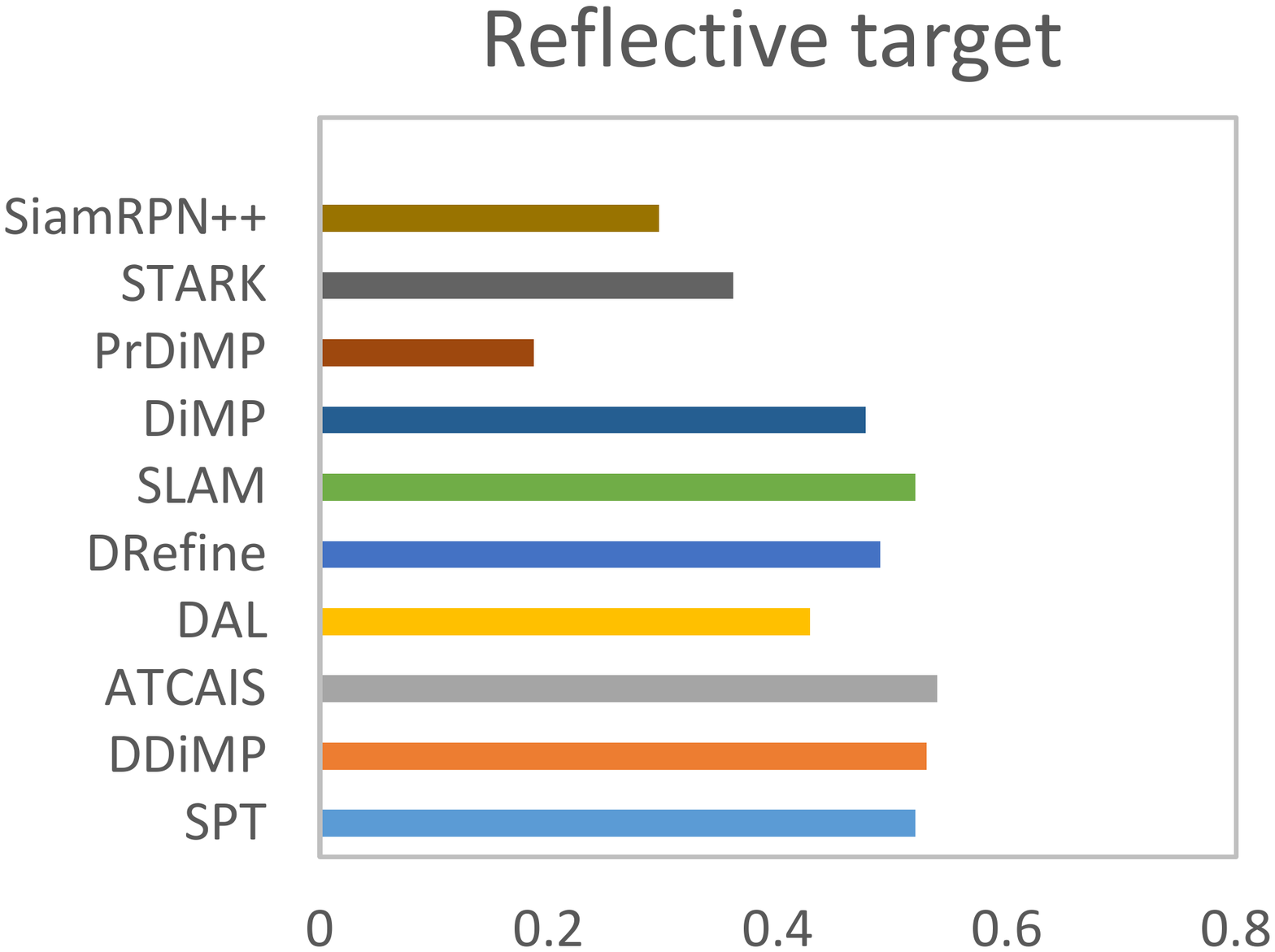}
\includegraphics[trim={46mm 26mm 52mm 30mm},clip,width=0.24\linewidth]{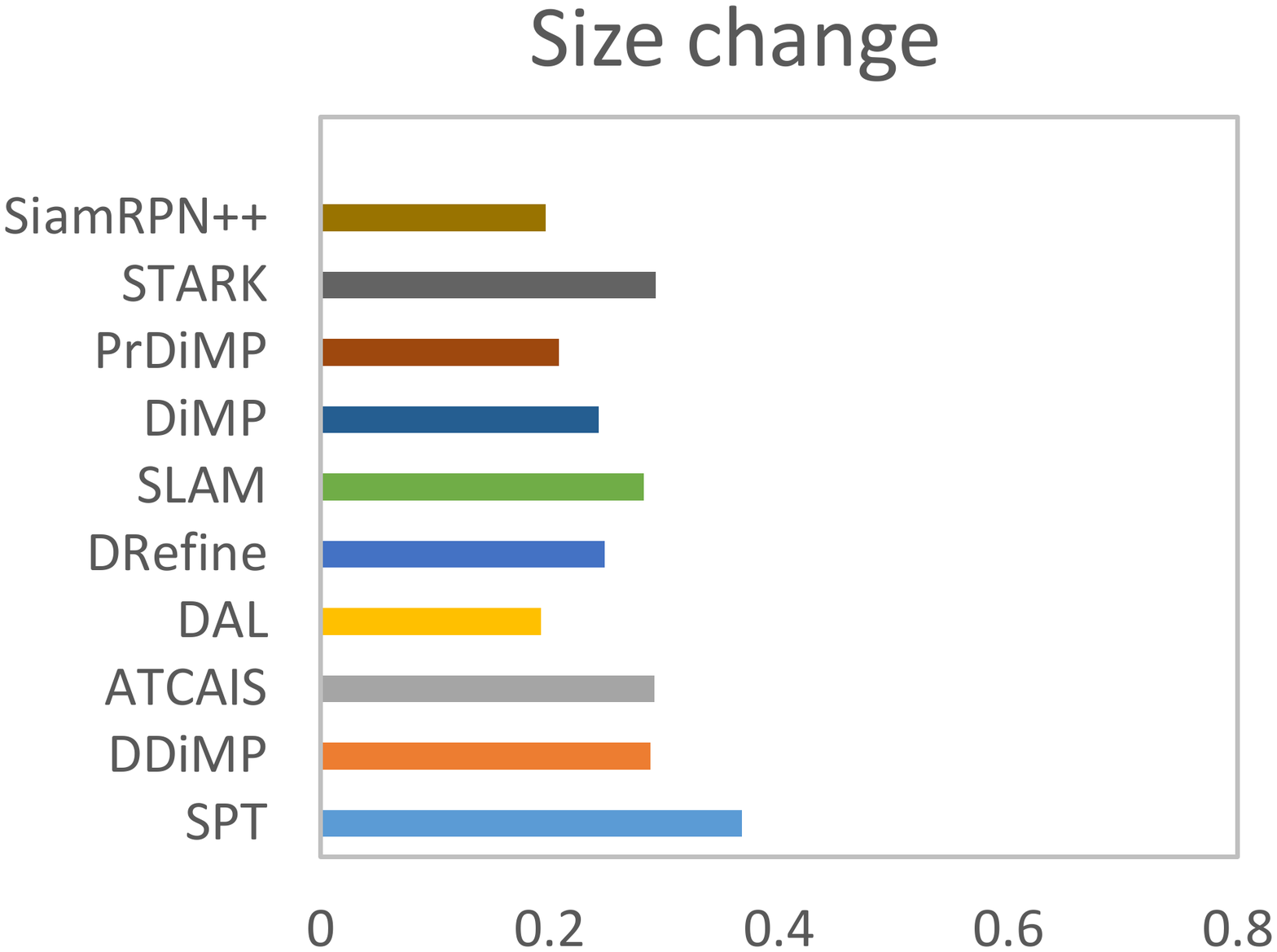}
\includegraphics[trim={46mm 26mm 52mm 30mm},clip,width=0.24\linewidth]{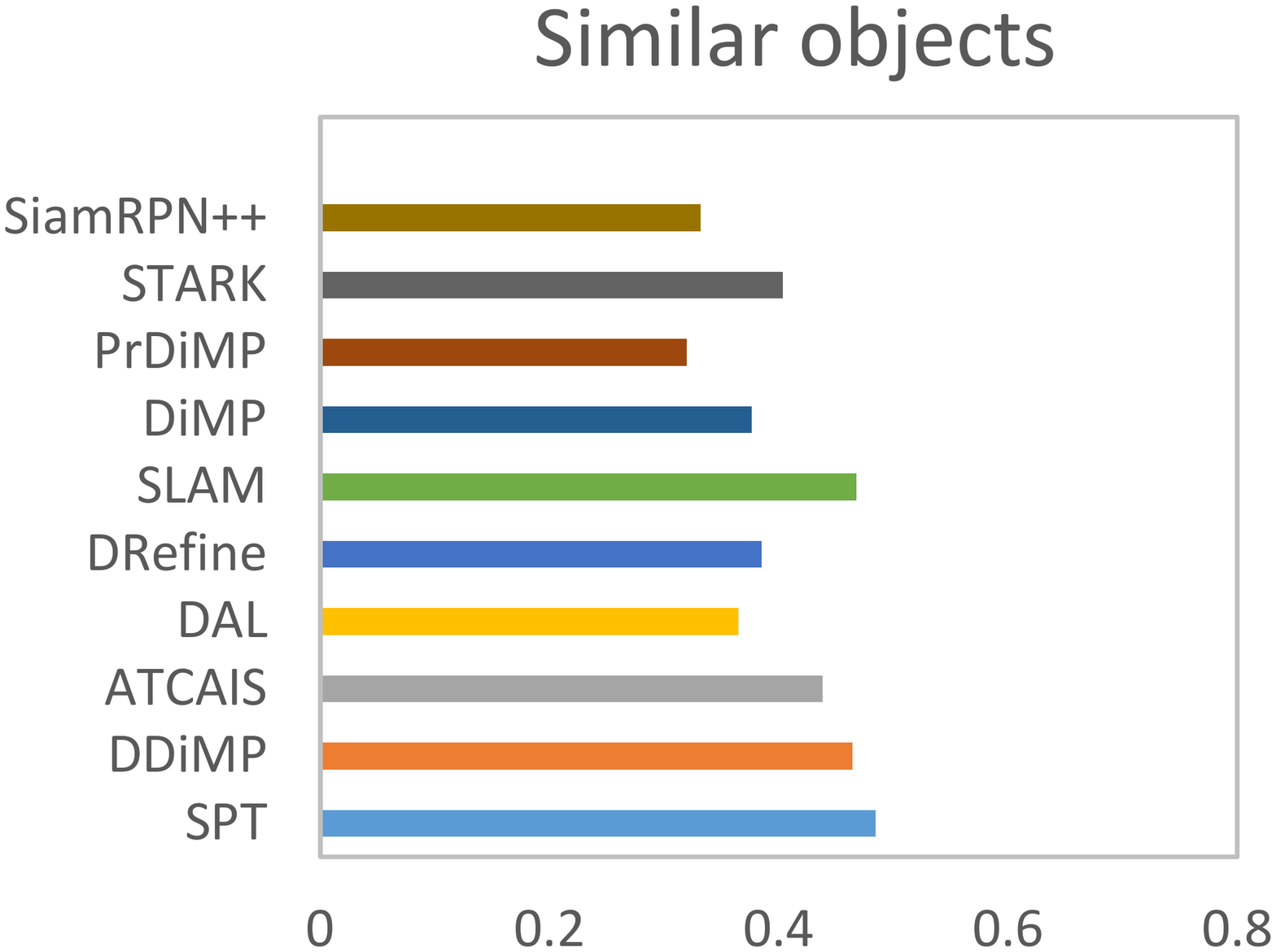}\\
\includegraphics[trim={46mm 26mm 52mm 30mm},clip,width=0.24\linewidth]{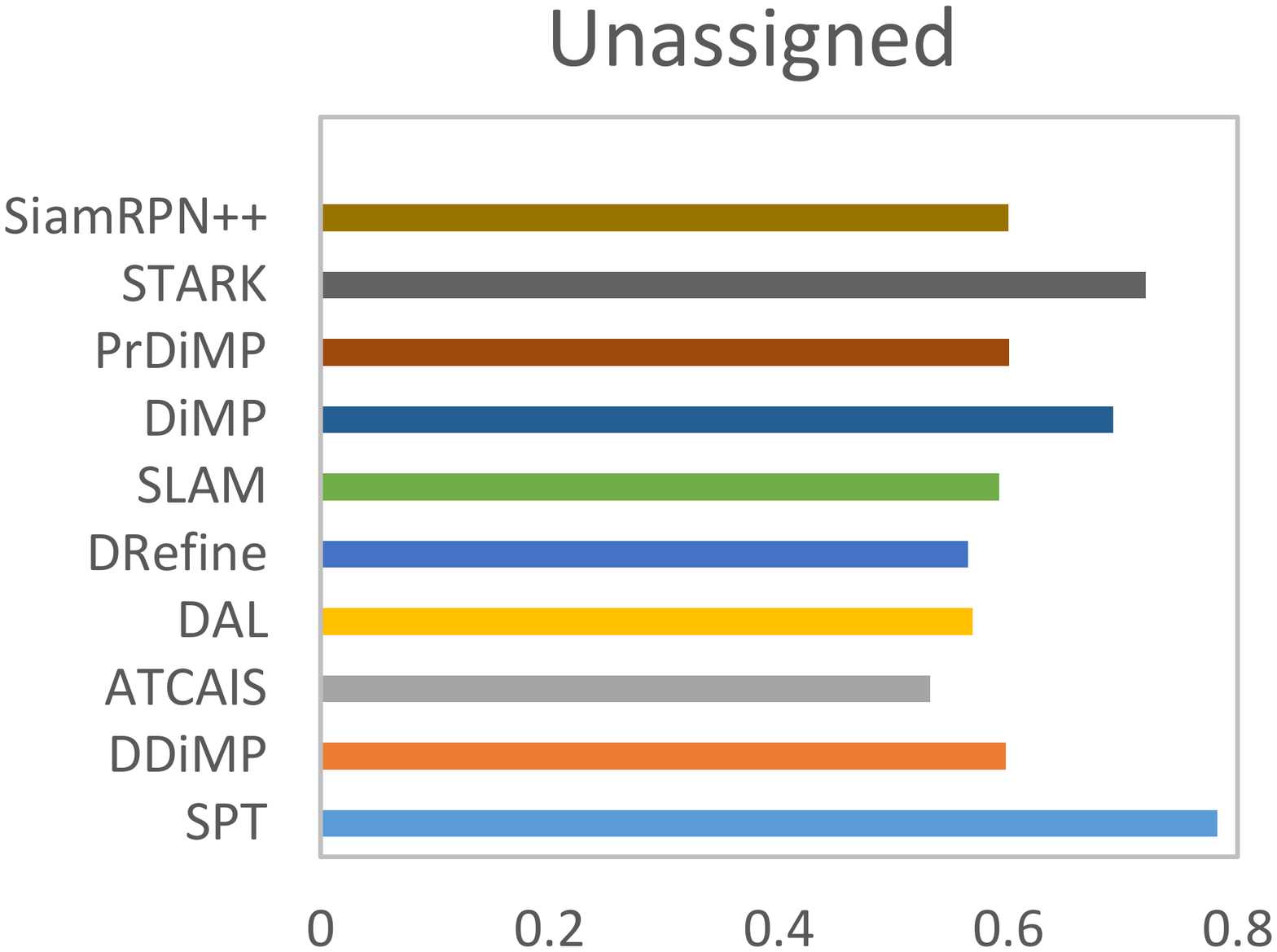}
\includegraphics[trim={46mm 26mm 52mm 30mm},clip,width=0.24\linewidth]{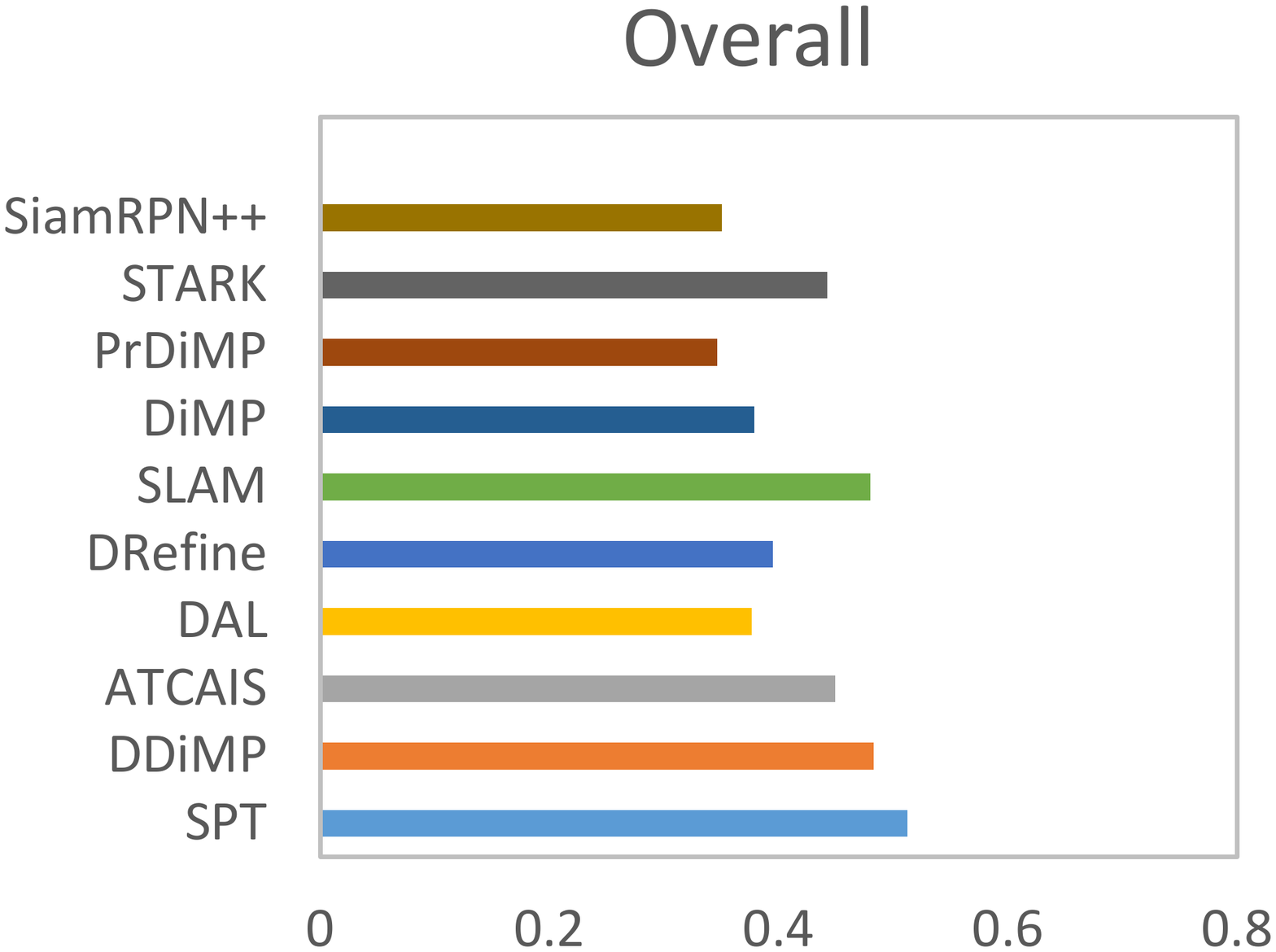}
\caption{Average overlap ratios of 10 trackers on the RGBD1K test set in terms of different challenging attributes. The horizontal axis shows the average overlap ratio of each tracker, and the vertical axis shows the algorithm names.}
\label{attribute_based}
\end{figure*}

\section{Experiments}
\textbf{Ablation study.} 
In terms of how to effectively fuse the features from two modalities, we conduct extensive experiments on both RGBD1K and CDTB datasets.
Above all, we construct four different structures of the fusion module, including Fusion A, Fusion B, Fusion C and Fusion D, as shown in Fig.~\ref{fusion_modules}. 
Fusion A only concatenates features from each modality across channels and introduces an $1d$ convolutional layer to reduce the channels number.
Fusion B introduces an additional transformer encoder to enhance the features based on the Fusion A structure.
Considering that the depth maps contain much noise that may damage the tracking performance, we try to emphasize the RGB modality by introducing a skip connection structure.
As shown in Fig.~\ref{fusion_modules}, Fusion C and Fusion D add the RGB features to the fused features based on Fusion A and Fusion B, respectively.

In Table~\ref{fuse_rgbd1k} and Table~\ref{fuse_cdtb}, we provide the results of four fusion strategies on the test set of RGBD1K and the CDTB dataset, respectively.
Comparing Fusion B with Fusion A, in terms of F-score, the 2-layer transformer encoder improves the performance by \emph{5.8\%} and \emph{1.9\%} on RGBD1K and CDTB, respectively.
However, when comparing Fusion C to Fusion A and Fusion D to Fusion B, respectively, the improvement from the skip connection is not pronounced, and sometimes may even bring negative effects.
To some extent, this phenomenon also reveals that in RGB-D tracking, depth channel presents comparable significance as RGB image, which should be dedicatedly modelled as the RGB image. 
Therefore, we select the Fusion B strategy in the proposed SPT tracker to fuse features from RGB and depth modalities.
Through a plain structure stacking only an $1d$ convolution layer and a 2-layer transformer encoder, compared with the results of STARK-S-FT tracker in the paper, Fusion B improves the F-score on RGBD1K and CDTB by \emph{7.5\%} and \emph{2.7\%}, respectively.
Besides, as a baseline method, the relatively pure structure and remarkable performance provide vigorous support for follow-up research in RGB-D tracking.

\textbf{Quantitative results.} 
We conduct a quantitative comparison of the proposed SPT with some state-of-the-art RGB-only trackers on the RGBD1K dataset.
Table~\ref{comparison} shows the results of trackers, including STARK~\cite{yan2021learning}, PrDiMP~\cite{danelljan2020probabilistic}, DiMP~\cite{bhat2019learning}, SiamRPN++~\cite{li2019siamrpn++}, KeepTrack~\cite{mayer2021learning}, KYS~\cite{bhat2020know}, TransT~\cite{chen2021transformer} and D3S~\cite{lukezic2020d3s}.
As can be seen in the table, the proposed SPT tracker outperforms other trackers in terms of Recall and F-score.
Besides, the SPT tracker achieves the second-best score on Precision.
\\\textbf{Qualitative results.}
In Fig.~\ref{qualitative_comparison1}, we provide a qualitative comparison of the tested RGB-D trackers, including SPT, DDiM~\cite{kristan2020eighth}, ATCAIS~\cite{kristan2020eighth}, SLAM~\cite{kristan2021ninth}, DRefine~\cite{kristan2021ninth} and DAL~\cite{qian2021dal}, on several challenging videos from the RGBD1K dataset.
As shown in the figure, although suffering from various challenging factors, such as background clutter, similar objects, partial occlusion, etc, the proposed SPT performs precise and steady tracking on these challenging sequences.

\textbf{Attribute-based analysis.} 
We report the attribute-based evaluation results of 10 trackers on the test set of RGBD1K in Fig.~\ref{attribute_based}.
The names of the trackers are displayed on the vertical axis.
We adopt the average overlap ratio between predicted bounding boxes and ground-truth bounding boxes as the metric for attribute-based evaluation.
Considering that there is no ground-truth bounding box when the target is full occluded (FO) or out of frame (OF), we only evaluate trackers on the remaining 13 attributes. 
The results show that the proposed tracker SPT outperforms other tackers in 8 attributes, including aspect change, background clutter, depth change, out-of-plane rotation, partial occlusion, similar object, size change and unassigned.
In other attributes, such as dark scene, non-rigid deformation and reflective target, the SPT is also competitive.
Evidently, learning from large-scale RGB and depth images, the SPT is more powerful to address various challenging factors causing drastic target appearance variations in RGB-D videos.

\end{document}